\documentclass[format=sigconf]{acmart}
\usepackage{graphicx}
\usepackage{booktabs} %
\usepackage[utf8]{inputenc} %
\usepackage[T1]{fontenc}    %
\usepackage{url}            %
\usepackage{booktabs}       %
\usepackage{amsfonts}       %
\usepackage{nicefrac}       %
\usepackage{microtype}      %
\usepackage{color}
\usepackage{subcaption}
\usepackage{enumitem}
\usepackage{xspace}
\usepackage[export]{adjustbox}
\usepackage{natbib}
\usepackage[printwatermark]{xwatermark}
\setcitestyle{square,numbers,sort}

\graphicspath{{figs/}}

\newif{\ifhidecomments}
\hidecommentsfalse
\ifhidecomments
    \newcommand{\chenhao}[1]{}
\else
    \newcommand{\chenhao}[1]{\textcolor{blue}{[#1 ---\textsc{ct}]}}
\fi

\ifhidecomments
    \newcommand{\viv}[1]{}
\else
    \newcommand{\viv}[1]{\textcolor{magenta}{[#1 ---\textsc{v}]}}
\fi

\newcommand{\spectrum}{spectrum\xspace}
\newcommand{\para}[1]{\noindent {\bf #1}\xspace}
\newcommand{\figref}[1]{Figure~\ref{#1}\xspace}
\newcommand{\secref}[1]{Section~\ref{#1}\xspace}

\begin{document}



\acmYear{2019}\copyrightyear{2019}
\setcopyright{acmcopyright}
\acmConference[FAT* '19]{FAT* '19: Conference on Fairness, Accountability, and Transparency}{January 29--31, 2019}{Atlanta, GA, USA}
\acmBooktitle{FAT* '19: Conference on Fairness, Accountability, and Transparency, January 29--31, 2019, Atlanta, GA, USA}
\acmPrice{15.00}
\acmDOI{10.1145/3287560.3287590}
\acmISBN{978-1-4503-6125-5/19/01}

\title{On Human Predictions with Explanations and Predictions of Machine Learning Models: A Case Study on Deception Detection}
\renewcommand{\shorttitle}{On Human Predictions with Explanations and Predictions of Machine Learning Models}
\author{Vivian Lai}
\affiliation{
  \institution{University of Colorado Boulder}
}
\email{vivian.lai@colorado.edu}

\author{Chenhao Tan}
\affiliation{
  \institution{University of Colorado Boulder}
}
\email{chenhao.tan@colorado.edu}

\begin{abstract}

Humans are the final decision makers in 
critical tasks that involve ethical and legal concerns, ranging from recidivism prediction, to medical diagnosis, to fighting against fake news.
Although machine learning models can 
sometimes 
achieve impressive performance in these tasks,
these tasks are {\em not} amenable to full automation.
To realize the potential of machine learning for improving human decisions, 
it is important to understand how
assistance from machine learning models affects human performance and human agency.

In this paper, we use deception detection as a testbed and investigate how we can harness explanations and predictions of machine learning models to improve human performance while retaining human agency.
We propose a \spectrum between full human agency and full automation, 
and 
develop varying levels of machine assistance along the \spectrum that gradually 
increase the influence of machine predictions.
We find that without showing predicted labels, explanations alone slightly improve human performance in the end task.
In comparison, human performance is greatly improved by showing predicted labels ($>$20\% relative improvement) and can be further improved by explicitly suggesting strong machine performance.
Interestingly, when predicted labels are shown, explanations of machine predictions 
induce a similar level of accuracy
as an explicit statement of strong machine performance.
Our results demonstrate a tradeoff between human performance and human agency and show that explanations of machine predictions can moderate this tradeoff.

\end{abstract}

\maketitle

\section{Introduction}
\label{sec:intro}

Machine learning has achieved impressive success in a wide variety of tasks.
For instance, 
neural networks have surpassed human-level performance in {\em ImageNet classification} (95.06\% vs. 94.9\%) \cite{he2015delving};
\citet{kleinberg2017human} demonstrate that in bail decisions, machine predictions of recidivism can reduce jail rates by 41.9\% with no increase in crime rates, compared to human judges;
\citet{ott2011finding} show that linear classifiers can achieve $\sim$90\% accuracy in detecting deceptive reviews while humans perform no better than chance.
As a result of these achievements, machine learning holds promise for addressing important societal challenges.
However, it is important to recognize different roles that machine learning can play in different tasks in the context of human decision making.
In tasks such as object recognition, human performance can be considered as the upper bound,
and machine learning models are designed to emulate the human ability 
to recognize objects in an image.
A high accuracy in such tasks presents great opportunities for large-scale automation and consequently improving our society's efficiency.
In contrast, efficiency is a lesser concern in tasks such as bail decisions.
In fact, full automation is often not desired in these tasks due to ethical and legal concerns.
These tasks are {\em challenging} for humans and for machines, but 
with vast amounts of data,
machines can sometimes identify patterns that are {\em unsalient, unknown, or counterintuitive} to humans.
If the patterns embedded in the machine learning models can be elucidated for humans, they 
can provide valuable support when humans make 
decisions.

\begin{figure*}[t]
  \includegraphics[width=0.95\textwidth]{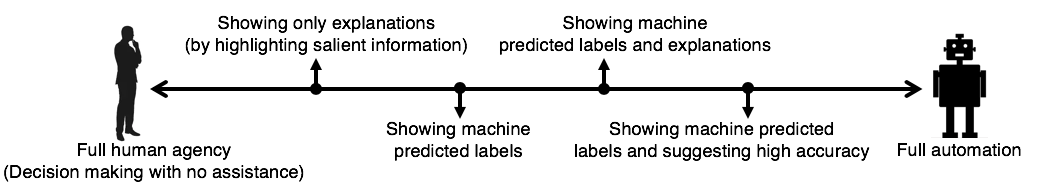}
  \caption{A \spectrum between full human agency and full automation illustrating how machine learning can be integrated in human decision making.
  The detailed explanation of each method 
  is in \secref{sec:approach}.}
  \label{fig:spectrum}
\end{figure*}

The goal of our work is to investigate best practices for integrating machine learning into human decision making.
We propose a \spectrum between full human agency, where humans make decisions entirely on their own,
and full automation, where machines make decisions without human intervention 
(see \figref{fig:spectrum} for an illustration).
We then develop varying levels of machine assistance along the \spectrum using explanations and predictions of machine learning models.
We build on recent developments in interpretable machine learning that provide useful frameworks for generating explanations of machine predictions \citep{kim2016examples,ribeiro2016should,lundberg2017unified,anchors:aaai18,lei2016rationalizing,kim2014bayesian}.
Instead of using these explanations to help users debug machine learning models, we 
incorporate the explanations as assistance for humans
to improve {\em human} performance
while retaining human agency in the decision making process.
Accordingly, we directly evaluate human performance in the end task through user studies.

In this work, we focus on a constrained form of decision making where humans make individual predictions.
Specifically, we ask humans to decide whether a hotel review is genuine or deceptive based on the text.
This prediction problem allows us to focus on 
the {\em integration} of machine learning into human predictions.
In comparison, prior work in decision theory and decision support systems  focuses on modeling preferences and utilities as well as building knowledge databases and representations to reason about complex decisions \citep{keen1978decision,berger2013statistical,newell1972human,horvitz1999principles,shim2002past}.
Moreover, 
since many policy decisions can be formulated as prediction problems \citep{kleinberg2015prediction},
understanding human predictions with assistance from machine learning models constitutes an important step towards 
empowering humans with machine learning in critical challenging tasks.

\para{Deception detection as a testbed.} In this work, we use deception detection as our testbed for three reasons.
First, deceptive information is prevalent on the Internet.
For instance, \citet{ott2012estimating} find that deceptive reviews are a growing problem on multiple platforms such as TripAdvisor and Yelp.
Fake news has also received significant attention recently \citep{lazer2018science,vosoughi2018spread} and might have influenced the outcome of the U.S. presidential election in 2016 \citep{allcott2017social}. 
Enhancing humans' ability in detecting deception can potentially alleviate these issues.

Second, deception detection is a challenging task for humans and has been extensively studied \citep{feng2012syntactic,feng2013detecting,ott2011finding,abouelenien2014deception,akoglu2013opinion}.
It is promising that machines show preliminary success in prior work.
For example,
machines are able to achieve an accuracy of $\sim$90\% in distinguishing genuine reviews from deceptive ones,
while human performance is no better than chance \citep{ott2011finding}.
Machines can identify unsalient and counterintuitive signals, 
e.g., deceptive reviews are less specific about spatial configurations and tend to include less sensorial and concrete language.
It is worth noting that we should take the high machine accuracy with a grain of salt in the general domain because deception detection is a complex problem.\footnote{For instance, one can argue that it is impossible to fully address the issue of deception in online reviews only based on textual information as an adversarial user can copy another user's review, which becomes a deceptive review but with exactly the same text as a genuine one.} 
The task introduced by \citet{ott2011finding} nevertheless 
provides an ideal sandbox
to %
understand human predictions
with 
assistance from machine learning models.

Third, full automation is not desired in critical tasks such as deception detection because of ethical and legal concerns.
The government should not have the authority to automatically block information from individuals, e.g., in the context of ``fake news''.
Furthermore, full automation may not comply with legal requirements. 
For instance, in the case of recidivism prediction, the Wisconsin Supreme Court ruled that ``judges be made aware of the limitations of risk assessment tools'' and ``a COMPAS risk assessment
should not be used to determine the severity of a sentence or
whether an offender is incarcerated'' \citep{nytimes,wicourts}.
Similarly, the trial judge is required to act as a gatekeeper regarding the evidence from a polygraph (lie detector)~\citep{daubert}.
Therefore, it is crucial to retain human agency 
and understand human predictions with assistance from machine learning models.

\para{Organization and Highlights.}
We start by reviewing related work to provide the necessary background for our study (\secref{sec:related}).
Our focus in this work is on
investigating human predictions with assistance from machine learning models in the context of deceptive review detection.
To explore the \spectrum between full human agency and full automation in \figref{fig:spectrum},
we develop varying levels of assistance from machine learning models (\secref{sec:approach}).
For example, the following three levels of machine assistance gradually increase the influence of machine predictions:
1) showing only explanations of machine predictions {\em without} revealing predicted labels;
2) showing predicted labels without revealing high machine accuracy;
3) showing predicted labels with an explicit statement of strong machine accuracy.
In \secref{sec:results}, we investigate human performance under different experimental setups along the \spectrum.
We show that explanations alone slightly improve human performance, while
showing predicted labels achieves great improvement ($\sim$21\% relative improvement in human accuracy).
However, this improvement is still moderate compared to ``full'' priming with an explicit statement of machine accuracy ($\sim$46\% relative improvement in human accuracy).
Our findings suggest that there exists a tradeoff between human performance and human agency.
Interestingly, when predicted labels are shown, explanations of machine predictions 
can achieve a similar effect as an explicit statement of machine accuracy.
We also find that humans tend to trust correct machine predictions more than incorrect ones, indicating that they can somewhat identify when machines are correct.

We further examine the effect of statements of machine accuracy by varying the accuracy numbers (\secref{sec:varying_accuracy}). 
Surprisingly, we find that our participants are not sensitive to statements of machine accuracy and are more likely 
to trust machine predictions with an accuracy statement than without, even if the accuracy statement suggests poor machine performance.
These observations echo with prior work on numeracy and suggest that it is difficult for humans to interpret and act on numbers \citep{peters2006numeracy,slovic2006risk,berwick1981doctors,reyna2008numeracy}.
We also find that frequency 
explanations (e.g., 5 out of 10 for explaining 50\%) can help 
humans calibrate the accuracy numbers.
Note that we do not recommend these presentations on the \spectrum because they present untruthful information.

We discuss the limitations of our work and provide concluding thoughts regarding future directions of investigating best practices for integrating artificial intelligence into human decision making in \secref{sec:conclusion}. 

\section{Related Work}
\label{sec:related} 

We summarize related work in two areas to put our work in context: interpretable machine learning and deception and misinformation.

\para{Interpretable machine learning.} 
Machine learning models remain as black boxes despite wide adoption.
Blindly following machine predictions may lead to dire repercussions, especially in scenarios such as medical diagnosis and justice systems \cite{caruana2015intelligible,kleinberg2017human,varshney2016engineering}.
Therefore, improving their transparency and interpretability has attracted broad interest \citep{kim2016examples,ribeiro2016should,lundberg2017unified,anchors:aaai18,lei2016rationalizing,kim2014bayesian}, dating back to early work on recommendation systems \citep{herlocker2000explaining,cosley2003seeing}.
In the case of general automation,
researchers have also studied issues of appropriate reliance and trust %
\citep{lee2004trust,wickens2015engineering,parasuraman1997humans,bussone2015role,dzindolet2003role}.

There are two major approaches to providing explanations of machine learning models: example-based and feature-based.
For example, an example-based explanation framework is MMD-critic proposed by \citet{kim2016examples}, which selects both prototypes and criticisms.
\citet{ribeiro2016should} propose a feature-based approach, LIME, that fits a sparse linear model to approximate non-linear models locally.
Similarly, \citet{lundberg2017unified} present a unified framework that assigns each feature an importance value for a particular prediction.

We would like to emphasize two unique aspects of our work: task difficulty and interpretability evaluation.
First, compared to categorizing text into topics and object recognition, deception detection is a challenging task for humans and it remains an open question whether humans can leverage help from machine learning models in such settings.
Second, we directly measure human performance in the end task. 
In comparison, prior work in interpretable machine learning aims to help humans understand how machine learning models work and/or debug them, the evaluation is thus mostly based on 
either the understanding of the models or the improvement in machine performance.
Concurrently, several recent studies have also examined how explanations relate to human performance~\citep{chandrasekaran2018explanations,feng2018can}.
Our work also resonates with the seminal work on mixed-initiative user interfaces \citep{horvitz1999principles} and intelligence augmentation \citep{ashby1957introduction}.
In addition, our work is connected to cognitive studies on understanding effective explanations beyond the context of machine learning \citep{lombrozo2007simplicity,lombrozo2006structure}.

\para{Deception and misinformation.}
Deception is a widely studied phenomenon in many disciplines~\citep{vrij2000detecting}.
In psychology, deception is defined as an act that is intended to foster in another person a belief or understanding which the deceiver considers false \citep{krauss1976modalities}.
To detect deception, researchers have examined the role of behavioral, emotional, and linguistic cues \citep{ekman1980facial,dulaney1982changes,knapp1974exploration,mehrabian1971silent,l1979telling,vrij2000detecting}.

Since people are increasingly relying on online reviews to make purchase decisions \citep{ye2011influence,zhang2010impact,chevalier2006effect,trusov2009effects}, machine learning methods have been used to detect deception in online reviews \citep{jindal2008opinion,wu2010distortion,yoo2009comparison,ott2011finding,feng2012syntactic,feng2013detecting}.
An important challenge in detecting deception in online reviews is to obtain the groundtruth labels of reviews.
\citet{ott2011finding} create the first sizable dataset in deception detection by asking Amazon Mechanical Turkers to write deceptive reviews.
Deceptive reviews can also be seen as an instance of spamming and online fraud~\citep{drucker1999support,gyongyi2004combating,ntoulas2006detecting,akoglu2013opinion}.

More recently, the issue of misinformation and fake news has drawn much attention from both the public and the research community \citep{farsetta2006fake,lazer2018science}.
Most relevant to our work is \citet{zhang2018structured}, which explores varying types of credibility annotations specifically designed for news articles.
In addition, \citet{nyhan2010corrections} demonstrate the ``backfire'' effect, which suggests that corrections of misperceptions may enhance people's false beliefs,
and \citet{vosoughi2018spread} show that fake news is more innovative and spreads faster than real news.

It is worth noting that deception detection is a broad and complex issue. For instance, 
fake news can be hard to define and may not be easily separated into two classes. 
Moreover, detecting fake news is different from detecting deceptive reviews as the former task requires other skills such as fact checking.
It is important to note that our focus in this work is on investigating {\em how humans interact with assistance from machine learning algorithms in decision making}.
We thus adopt the task of distinguishing genuine reviews from deceptive ones based on textual information in \citet{ott2011finding} as a sandbox.
Our results on this constrained deception detection task can potentially contribute valuable insights to future solutions of the broader issue of deception detection.

\section{Experimental Setup and Hypotheses}
\label{sec:approach}

Our goal is to understand whether  machine predictions and their explanations can improve human performance in challenging tasks, such as deception detection, and how humans interpret assistance from machine learning models.
In this section, we first present our task setup and then develop 
varying levels of machine assistance along the \spectrum introduced in \figref{fig:spectrum}.
We finally formulate our hypotheses and define our evaluation metrics.

\begin{figure*}[t]
\centering
\begin{subfigure}[t]{\textwidth}
  \centering
  \includegraphics[width=0.85\textwidth,frame]{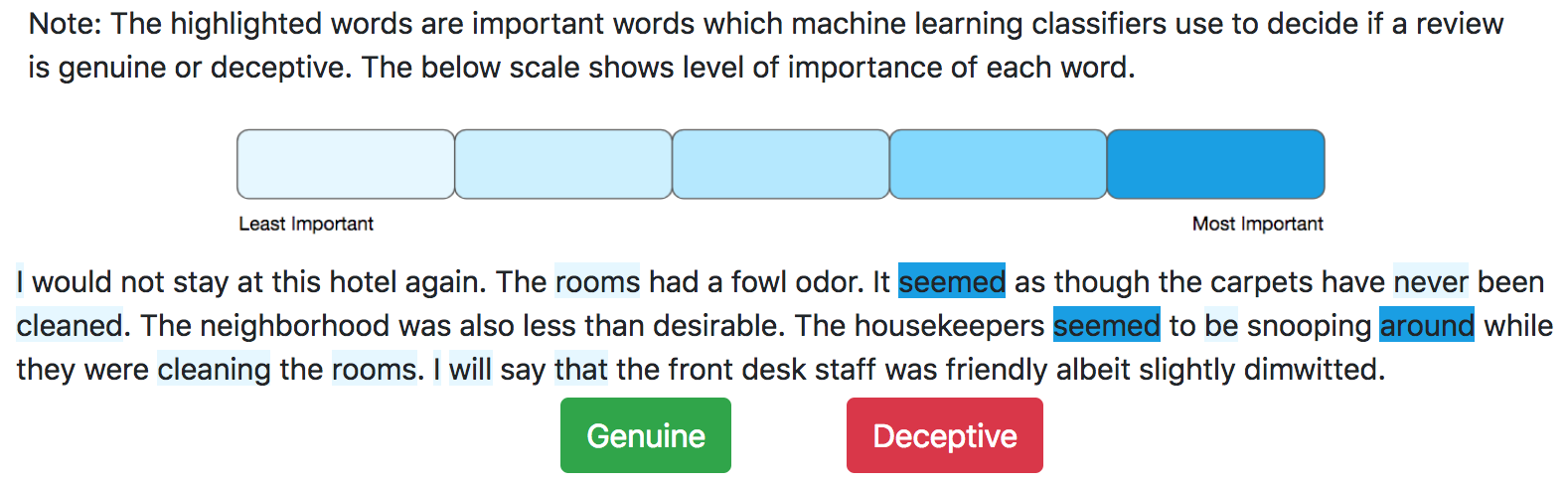}
  \caption{Heatmap (without showing predicted labels), an instance of feature-based explanations.}
  \label{fig:exp-heatmap}
\end{subfigure}
\begin{subfigure}[t]{\textwidth}
  \smallskip
  \centering
  \includegraphics[width=0.85\textwidth,frame]{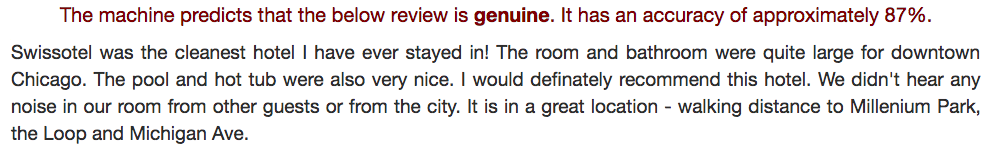} 
  \caption{Predicted label with accuracy.}
  \label{fig:exp-w-acc}
\end{subfigure}
\begin{subfigure}[t]{\textwidth}
  \centering
  \smallskip
  \includegraphics[width=0.85\textwidth,frame]{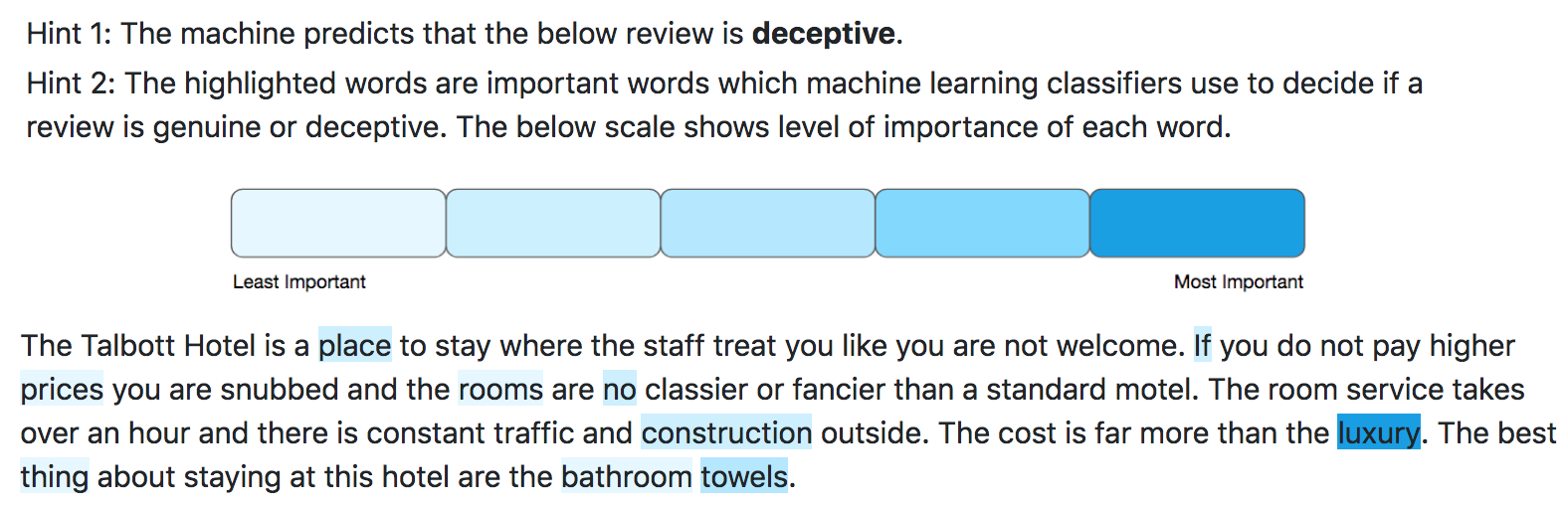} 
  \caption{Predicted label + heatmap (without accuracy).}
  \label{fig:exp-comb1}
\end{subfigure}
\caption{Example interfaces with varying levels of machine assistance.
\figref{fig:exp-heatmap} only presents feature-based explanations of machine predictions in the form of {\em heatmap}.
\figref{fig:exp-w-acc} shows both the predicted label and an explicit statement about machine accuracy (87\%).
\figref{fig:exp-comb1} shows the predicted label with heatmap, but does not present machine accuracy.
We crop the ``Genuine'' and ``Deceptive'' buttons in Figure~\ref{fig:exp-w-acc} and \ref{fig:exp-comb1} to save space.}
\label{fig:exp-interfaces}
\end{figure*}

\para{Experimental setup.}
We employ the deception detection task developed by \citet{ott2011finding} and evaluate human performance in this task with varying levels of machine assistance.
The dataset in \citet{ott2011finding} includes 800 genuine and 800 deceptive  hotel reviews for 20 hotels in Chicago.
The genuine reviews were extracted from TripAdvisor and the deceptive ones were written by turkers.
We use 80\% of the reviews as training data and the remaining 20\% as the heldout test set. 
Since the machine performance with linear SVM in \citet{ott2011finding} already surpasses humans ($\sim$50\%) by a wide margin and linear classifiers are generally considered more interpretable, we follow \citet{ott2013negative} and use linear SVM with bag-of-words features as our machine learning model.
The linear SVM classifier achieves an accuracy of 87\% on the heldout test set.

Our main task in this paper is to evaluate human performance with assistance from machine learning models.
To do that, we conduct a user study on Amazon Mechanical Turk.
Turkers are recruited to determine whether a review in the heldout test set is genuine or deceptive.
In other words, humans are asked to perform the same task as the \textit{machine} on the test set.
We follow a between-subject design:
each turker 
is assigned a level of machine assistance along the spectrum (\figref{fig:spectrum}) and 
labels 20 reviews after going through three training examples and correctly answering an attention-check question.
To incentivize turkers to perform at their best, we provide 40\% bonus for each correct prediction in addition to the 5 cent base rate for a review.
We also solicit our participants' estimation of their own performance and basic demographic information such as gender and education background through an exit survey.
We only allow a turker to participate in the study once to guarantee sample independence across experimental setups.
Given that there are 320 test reviews and that we collect five turker predictions for each review, 
each experimental setup has a total of 80 turkers.
Refer to the appendix for more details regarding our user study and survey questions.

\begin{figure*}
\centering
\begin{subfigure}[t]{.49\textwidth}
  \centering
  \includegraphics[width=0.84\textwidth]{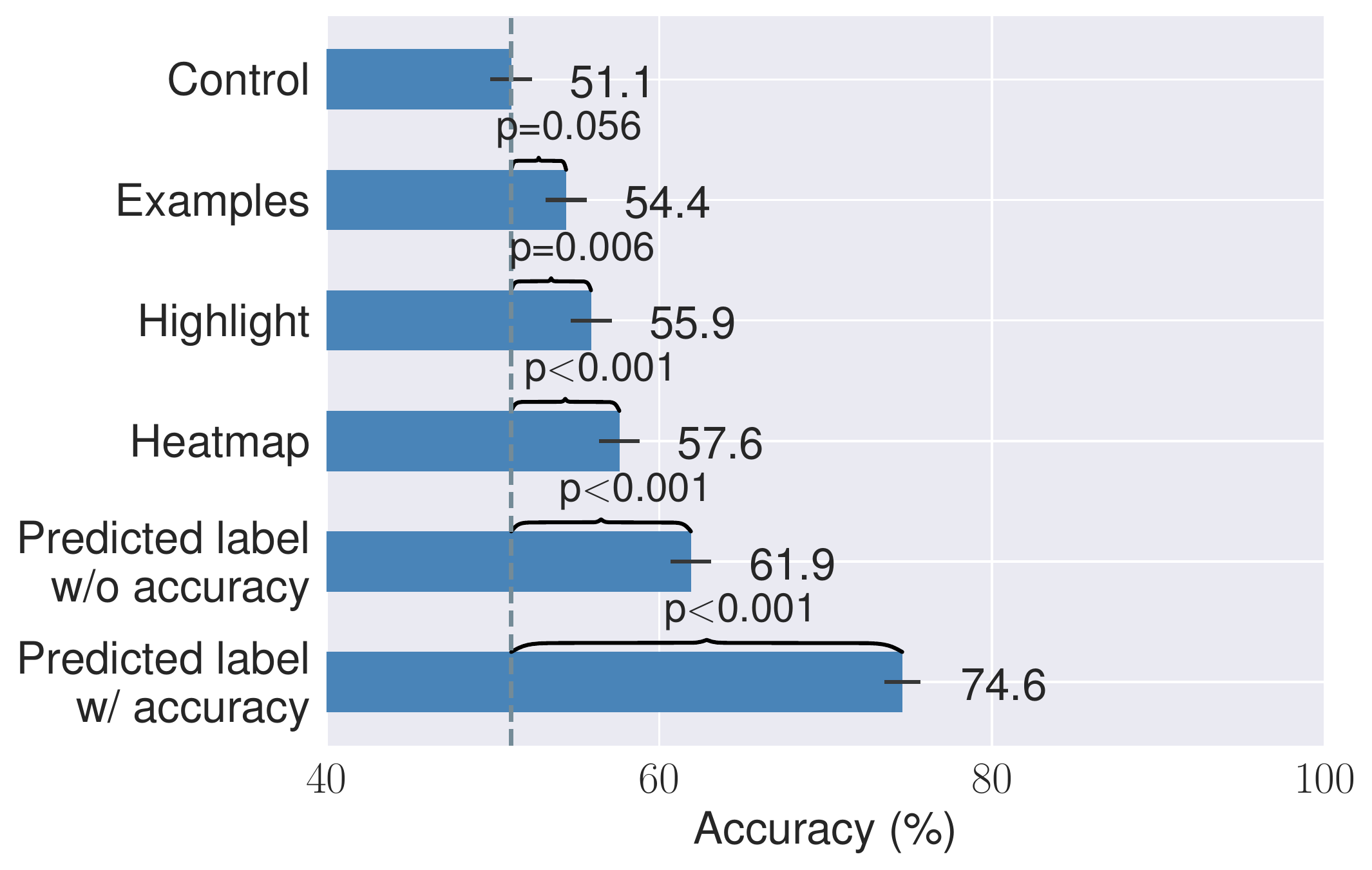}
  \caption{Human accuracy with varying levels of machine assistance.}
  \label{fig:exp-machine-assist}
\end{subfigure}
\hfill
\begin{subfigure}[t]{.49\textwidth}
  \centering
  \includegraphics[width=0.84\textwidth]{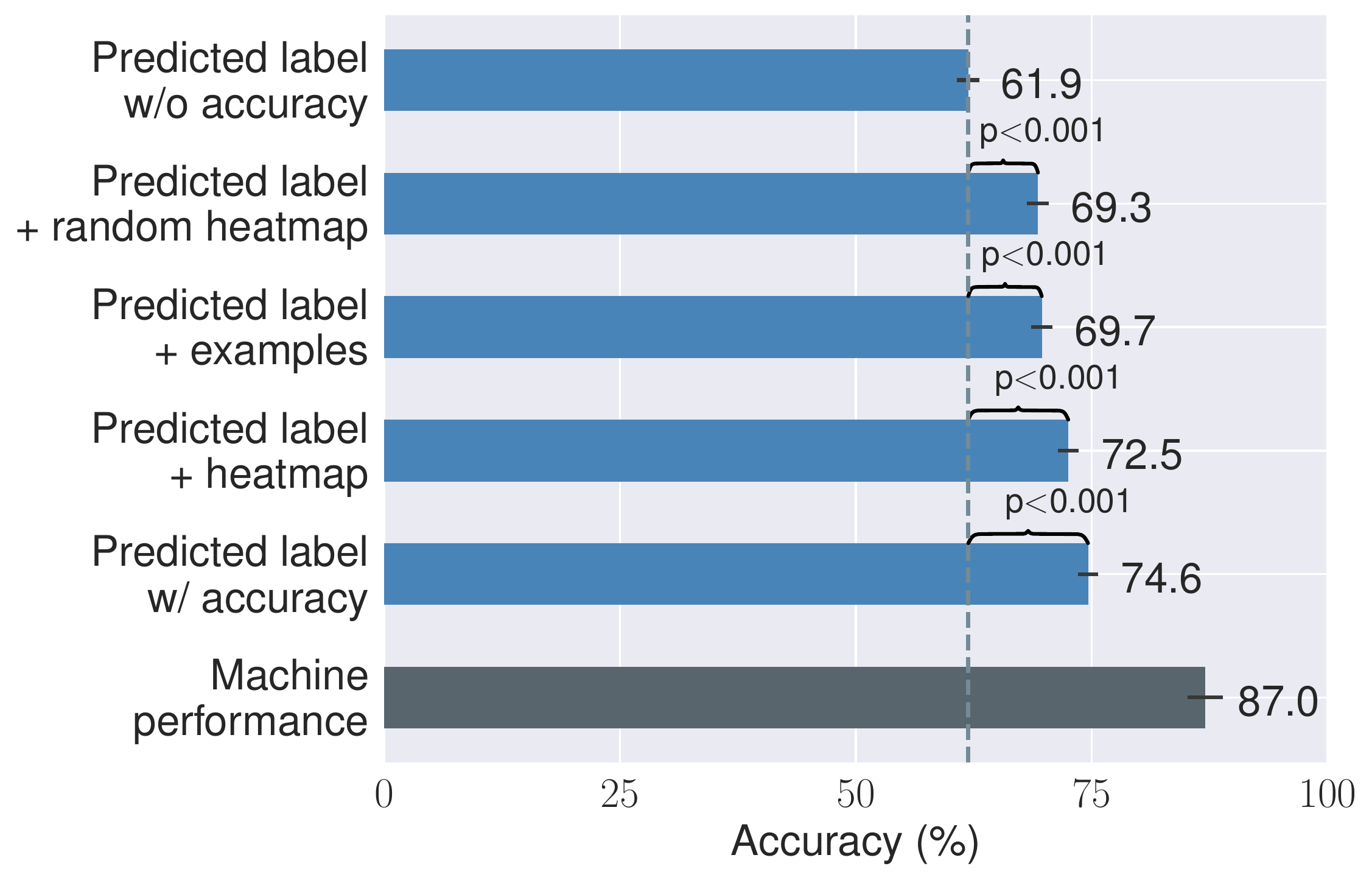}
  \caption{
  Human accuracy with predicted labels (and other information).
  }
  \label{fig:exp-machine-pred}
\end{subfigure}
\caption{
Human accuracy with varying levels of assistance.
In \figref{fig:exp-machine-assist},
{\em control} provides no assistance; {\em examples}, {\em highlight}, and {\em heatmap} present explanations of machine predictions alone;
{\em predicted label w/o accuracy} shows predicted labels;
{\em predicted label w/ accuracy} shows predicted labels and reports machine accuracy that suggests strong machine performance.
It is clear that showing predicted labels is crucial for improving human accuracy.
Adding an explicit statement of machine accuracy further improves human accuracy.
\figref{fig:exp-machine-pred} further investigates the combinations of predicted labels and their explanations, and presents
{\em machine performance} as a benchmark.
Intriguingly, we find that adding explanations achieves a similar effect as adding an explicit statement of machine accuracy.
All p-values are computed by conducting t-test between the corresponding setup and the first experimental setup in the figure (``control'' in \figref{fig:exp-machine-assist} and ``predicted label w/o accuracy'' in \figref{fig:exp-machine-pred}).
}
\label{fig:exp-machine}
\end{figure*}

\para{Varying levels of machine assistance.}
Humans are the main agents in our experiments and make final decisions; machines only provide assistance, which can be ignored if humans deem it useless. 
An ideal outcome is that human performance can be improved with minimal information from machine learning models so that humans retain their agency in the decision making process.
To examine how humans perform under different levels of influence from machine learning models,
we consider the following presentations 
along the \spectrum in \figref{fig:spectrum} (we only show three interfaces in \figref{fig:exp-interfaces}
for space reasons; see the appendix for more).

\begin{itemize}[topsep=0pt,itemsep=0pt,leftmargin=*]
  \item {\bf Control.} Humans are only presented a review. This setup contains no information from machine learning models and humans have full agency.
  
  \item {\bf Feature-based explanations.} Since our machine learning model is linear, we present two versions of feature-based explanations by highlighting words based on absolute values of weight coefficients. 
  First, we highlight the top 10 words in each review with the same color ({\em highlight}).
  Second, we use {\em heatmap} to show gradual changes in weight coefficients among the top 10 words.
  The most heavily-weighted words are highlighted in the darkest shade of blue.
  Soft-highlighting (heatmap) has been shown to improve visual search on targeted areas for humans \citep{kneusel2017improving}.
  Note that we do not indicate the sign of features to avoid revealing predicted labels.
  Humans may pay extra attention to the highlighted words and accordingly make decisions on their own.
  Figure \ref{fig:exp-heatmap} shows an example interface for {\em heatmap}.

  \item {\bf Example-based explanations.} This method ({\em examples}) is inspired by example-based interpretable machine learning \cite{kim2016examples}. Humans are presented two additional reviews from the training data, one deceptive and one genuine that are most similar to the review under consideration. This setup resonates with nearest neighbor classifiers. 
  Humans can potentially make better decisions in this setup than in {\em control} by comparing the similarity between reviews.

  \item {\bf Predicted label without accuracy.} The above two approaches only show explanations of machine predictions, but do not reveal any information about predicted labels. 
  The next level of priming presents the predicted label.
  If humans fully follow machine predictions, they will perform much better than chance and likely lead to an upper bound in this deception detection task for humans. 
  However, humans may not trust the machine due to algorithm aversion \citep{dietvorst2015algorithm}.

  \item {\bf Predicted label with accuracy.} We may further influence human decisions by explicitly suggesting that machines perform well in this task with 87\% accuracy.
  Figure \ref{fig:exp-w-acc} shows an example for {\em predicted label with accuracy}.
  Note that such strong recommendations may not be desired due to ethical and legal concerns (see our discussion in the introduction).
  
  \item {\bf Combinations.} Finally, we combine feature (example)-based explanations and predicted labels.
  Note that we do not show machine performance to avoid strong priming.
  Figure \ref{fig:exp-comb1} shows an example of {\em predicted label + heatmap} without information about machine performance.
\end{itemize}

\para{Hypotheses.}
We formulate the following hypotheses 
regarding {\em how well humans can perform with machine assistance}
and {\em how often humans trust machine predictions when predicted labels are available}.

\begin{itemize}[topsep=0pt,itemsep=0pt,leftmargin=*]
 \item \textit{Hypothesis 1a.} Feature-based explanations and example-based explanations improve human performance over {\em control}.
 \item \textit{Hypothesis 1b.} {\em Heatmap} is more effective than {\em highlight} as gradual changes in weight coefficients can be useful, as shown in \citet{kneusel2017improving} for visual search.
 Feature-based explanations are more effective than example-based explanations since the latter requires a greater cognitive load, i.e., reading two more reviews.
  \item \textit{Hypothesis 2.} {\em Showing predicted labels} significantly improves human performance compared to feature (example)-based explanations alone.
  Assuming that humans trust the machine and follow its prediction, 
  showing predicted labels can likely improve human performance because the machine accuracy is 87\%.
  However, showing predicted labels 
  reduces human agency, so 
  it is important to understand
  the size of the performance gap and make informed design choices.
  \item \textit{Hypothesis 3.} 
  By combining predicted labels and feature (example)-based explanations, the trust that humans place on machine predictions increases, as it has been shown that concrete details can influence the level of trust in general automation \citep{lee2004trust}.
\end{itemize}

We evaluate the above hypotheses using two metrics, accuracy and trust.
{\em Accuracy} is defined as the percentage of correctly predicted instances by humans; {\em trust} is defined as the percentage of instances for which humans follow the machine prediction.
Note that we can only compute trust when predicted labels are available.

\section{Results}
\label{sec:results}

In this section, we investigate how varying levels of assistance from machine learning models along the \spectrum in \figref{fig:spectrum} affect human predictions.
We first discuss aggregate human performance using human accuracy and trust.
Our results show that in this challenging task, explanations alone 
slightly improve human performance, while showing predicted labels can significantly improve human performance.
When predicted labels are shown, we examine the level of trust that humans place on machine predictions.
Our results suggest that humans can somewhat differentiate correct machine predictions from incorrect ones.
Finally, we present individual differences among our participants based on information collected in the exit survey.
Our dataset and demonstration are available at \url{https://deception.machineintheloop.com/}.

\begin{figure*}
\centering
\begin{subfigure}[t]{0.43\textwidth}
  \centering
  \includegraphics[width=0.87\textwidth]{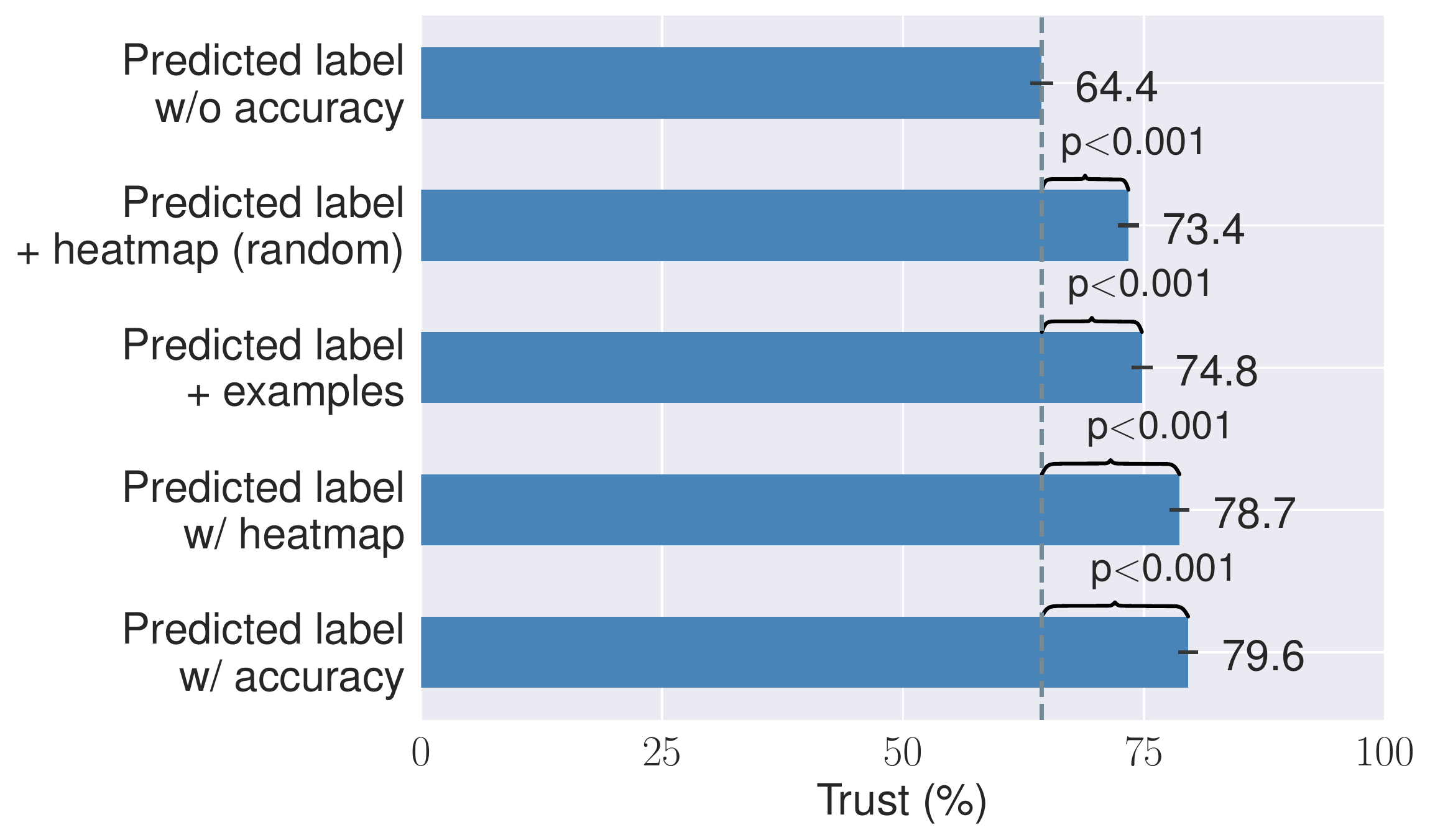}
  \caption{Trust in machine predictions.}
  \label{fig:exp-trust}
\end{subfigure}
\hfill
\begin{subfigure}[t]{.55\textwidth}
  \centering
  \includegraphics[width=0.86\textwidth]{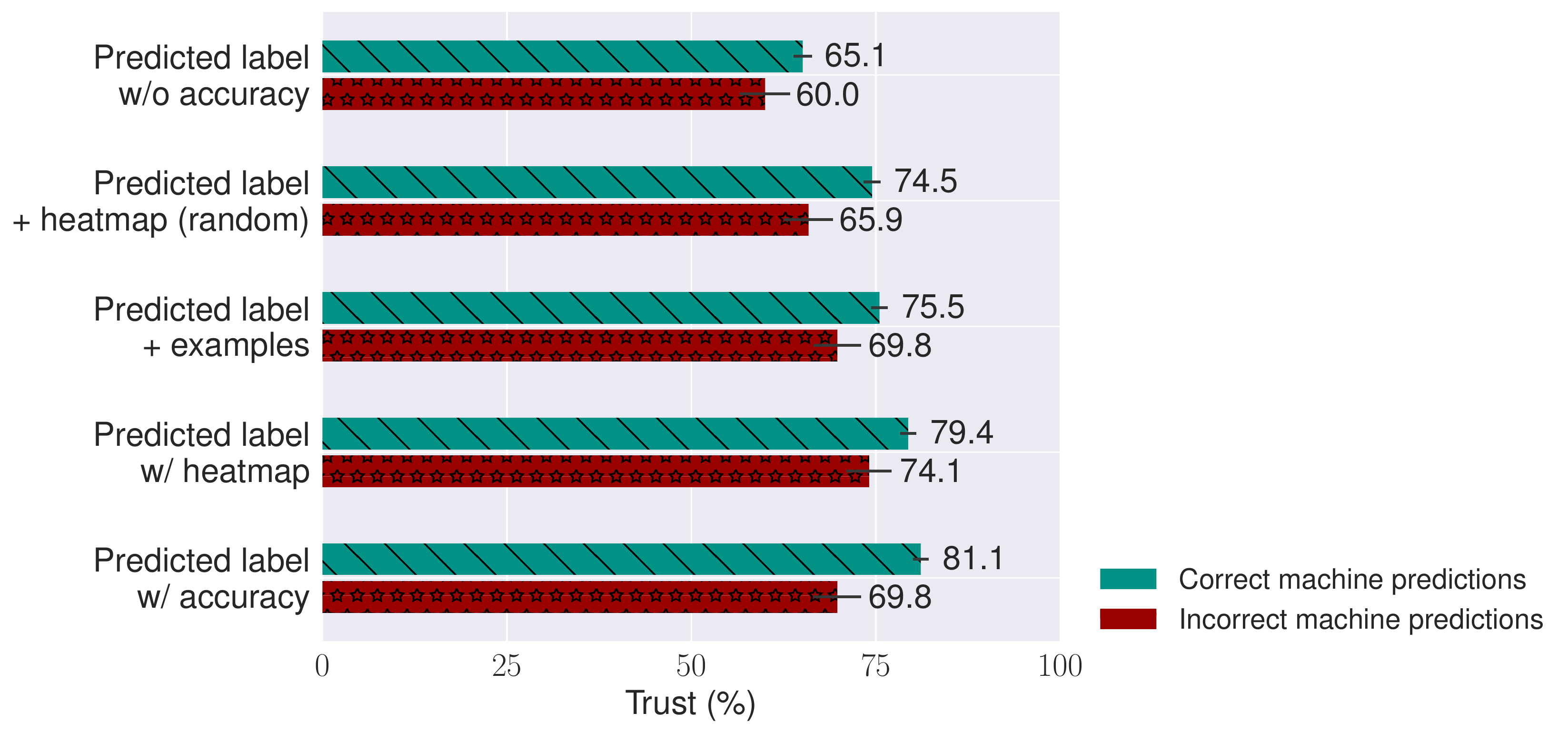}
  \caption{Trust in correct and incorrect machine predictions.}
  \label{fig:exp-trust-predictions}
\end{subfigure}
\caption{
The trust that humans place on machine predictions.
\figref{fig:exp-trust} shows that adding feature-based explanations (heatmap) can effectively increase the trust level compared to {\em predicted label w/o accuracy}.
$p$-value in \figref{fig:exp-trust} is computed by conducting t-test between the corresponding setup and {\em predicted label w/o accuracy}.
\figref{fig:exp-trust-predictions} breaks down the trust based on whether machine predictions are correct or incorrect and shows that humans trust correct machine predictions more than the incorrect ones in all the five experimental setups, although the differences are only statistically significant in two setups. 
}
\end{figure*}

\subsection{Human Accuracy}

We first present 
human accuracy measured by the percentage of correctly predicted instances by humans.
Our results suggest that showing predicted labels is crucial for improving human performance.
Featured-based explanations coupled with predicted labels are able to induce similar human performance as an explicit statement of strong machine accuracy.
As such, adding feature-based explanations to predicted labels may be more ideal than suggesting strong machine performance as the priming is weaker and may facilitate a higher level of human agency in decision making.

\para{Explanations alone slightly improve human performance (\figref{fig:exp-machine-assist}).}
As \figref{fig:exp-machine-assist} shows, human performance in {\em control} is no better than chance (51.1\%).
This finding is consistent with \citet{ott2011finding} and decades of research on deception detection \cite{bond2006accuracy}.
Explanations alone slightly improve human performance over control, and the differences are statistically significant for {\em highlight} and {\em heatmap}, not for {\em examples}.
However, the best explanations, {\em heatmap}, is not statistically significantly different from {\em highlight} ($p=0.335$) or {\em examples} ($p=0.069$).
As a result, our findings partially supports \textit{Hypothesis 1a} and rejects \textit{Hypothesis 1b}.

These findings suggest that it is difficult for humans to understand explanations on their own.
This is plausible for example-based explanations since it requires extra cognitive burden and estimating text similarity is a nontrivial task for humans.
For feature-based explanations, it seems that the improvement is driven by the small number of training reviews that we provide to explain the task.
First-person singular pronouns provide a good example: one of the training reviews is deceptive and highlight many occurrences of the word, ``my''.
A participant said,
{\em ``I tried to match the pattern from the example. In the example. the review with the most "My's" and "I's" were deceptive''}.
In other words, the improvement in {\em heatmap} and {\em highlight} may not happen at all without the training reviews, which indicates the difficulty of interpreting these feature-based explanations and the importance of explaining the explanations.
One possible direction is to develop automatic tutorials to teach the intuitions behind important features, which is related to machine teaching \citep{zhu2015machine,mac2018teaching,singla2014near}.

\para{Showing predicted labels significantly improves human performance
(\figref{fig:exp-machine-assist} and \ref{fig:exp-machine-pred}).}
As \figref{fig:exp-machine-assist} shows, showing predicted labels drastically improves human performance (61.9\% for {\em predicted label w/o accuracy}, a 21\% relative improvement over {\em control}; the difference with {\em heatmap} is statistically significant ($p<$0.001)).
By presenting machine accuracy as shown in \figref{fig:exp-w-acc}, the performance is further improved to 74.6\% ({\em predicted label w/ accuracy} in \figref{fig:exp-machine-assist}, 
a 46\% relative improvement over {\em control}).

These results are consistent with \textit{Hypothesis~2}.
The big performance gap between showing predicted labels and showing feature (example)-based explanations alone suggests that when humans interact with machine learning models, it makes a significant difference whether predicted labels are shown.
However, this observation also echoes with concerns about humans overly relying on machines \cite{lee2004trust}.

To further understand human performance with predicted labels, we examine all experimental setups with predicted labels in \figref{fig:exp-machine-pred}.
Although showing predicted labels seems necessary for achieving sizable human performance improvement,
the effect of presenting machine accuracy can be moderated by showing feature (example)-based explanations.
We find that {\em predicted label + examples} and {\em predicted label + heatmap} outperform {\em predicted label w/o accuracy} (69.7\% and 72.5\% vs. 61.9\%), without 
presenting the machine accuracy.
In this case, we observe that heatmap is more effective than examples, and leads to comparable human performance with {\em predicted label w/ accuracy}.
There is still a gap between the best human performance ({\em predicted label w/ accuracy}) and {\em machine performance} (74.6\% vs. 87.0\%).
These observations suggest that humans do not necessarily trust machine predictions.

\begin{figure*}
\centering
\begin{subfigure}[t]{0.53\textwidth}
  \centering
  \includegraphics[width=0.9\textwidth]{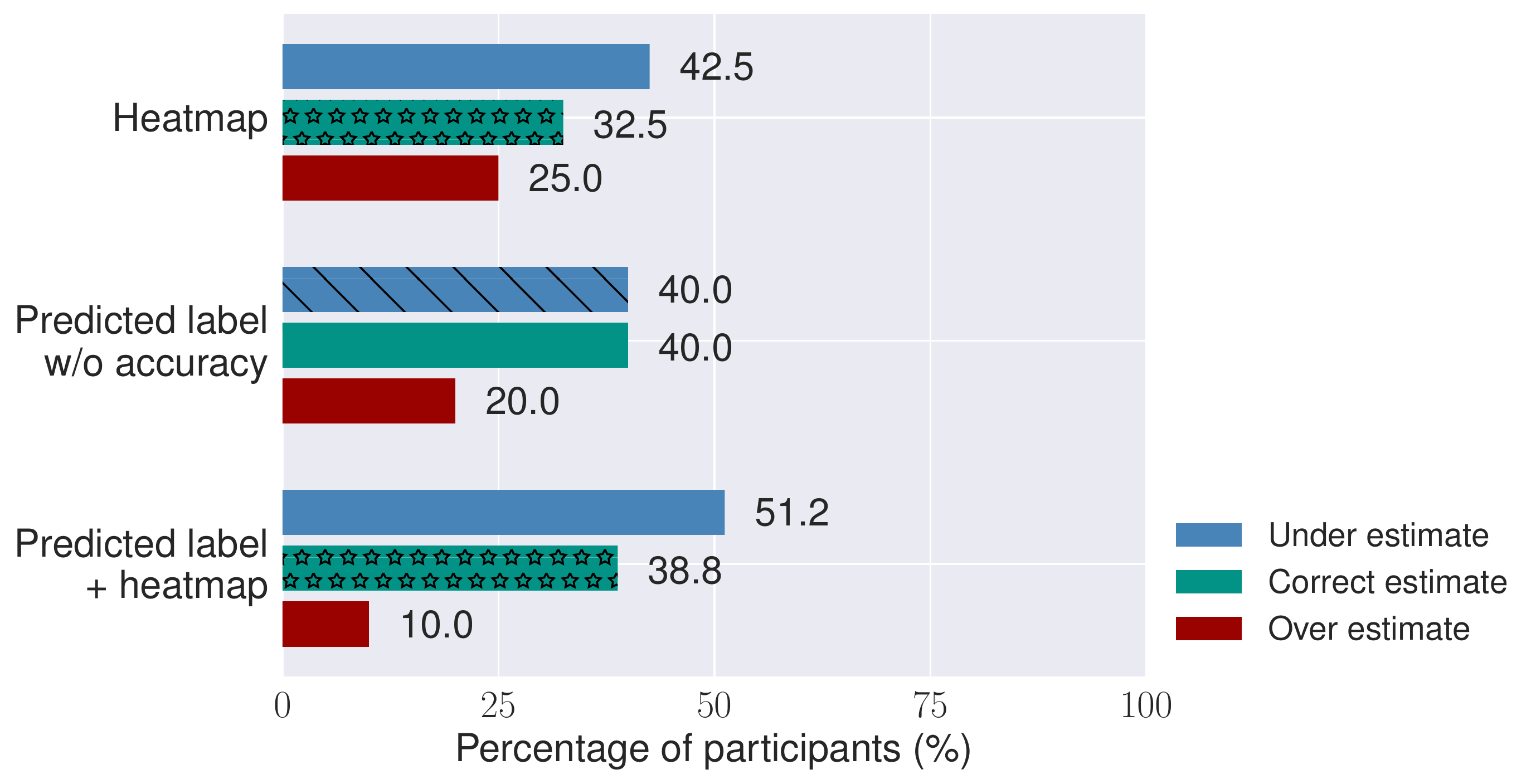}
  \caption{Human estimation of their own performance.}
  \label{fig:exp-estimate}
\end{subfigure}
\begin{subfigure}[t]{.45\textwidth}
  \centering
  \includegraphics[width=0.85\textwidth]{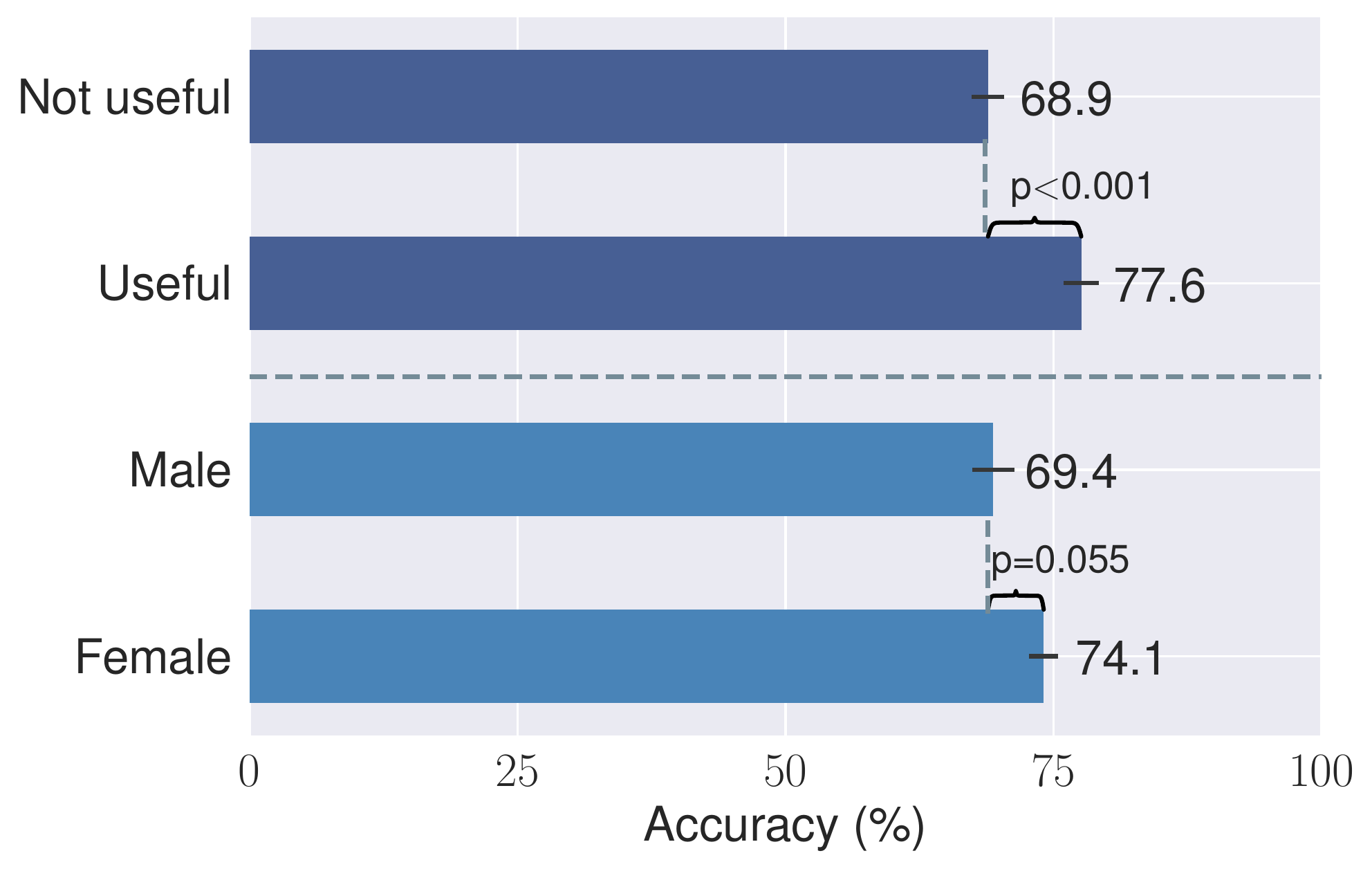}
  \caption{Gender and hint usefulness in {\em predicted label + heatmap}.}
  \label{fig:exp-demo}
\end{subfigure}
\caption{
Heterogeneity findings among participants in our study. Figure \ref{fig:exp-estimate} shows 
performance estimation by participants in three different experimental setups. Figure \ref{fig:exp-demo} presents the performance of participants in 
{\em predicted label + heatmap} group by two variables, hint usefulness and gender.
}
\end{figure*}

\subsection{Trust}

We further examine the levels of trust that humans place on machine predictions when predicted labels are available.
Since machine performance surpasses human performance in {\em control} by a wide margin in this task, higher levels of trust are correlated with higher levels of accuracy in our experiments.
However, these two metrics capture different dimensions of human predictions because trust is 
tied to machine predictions.
This becomes clear when
we break down human trust by whether machine predictions are correct or not.
We find that humans tend to trust correct machine predictions more 
than incorrect ones, which suggests that humans can somewhat effectively identify cases where machines are wrong.
It is important to emphasize that our focus is on understanding how human trust varies along the \spectrum rather than manipulating the trust of humans in machines.

\para{Feature (example)-based explanations increase the trust that humans place on machine predictions (\figref{fig:exp-trust}).}
We further introduce random heatmap by randomly highlighting an equal number of words as in {\em heatmap} to examine whether humans are influenced by any 
explanations
including random ones.

Our results are consistent with {\em Hypothesis 3}: both feature-based and example-based explanations increase the trust of humans in machine predictions.
In fact, {\em predicted label + heatmap} leads to a similar level of trust as {\em predicted label w/ accuracy}, although the latter explicitly tells humans that the machine learning model ``has an accuracy of approximately 87\%''.
In other words, when predicted labels are shown, heatmap can nudge humans in decision making without making strong statements of machine accuracy.
Interestingly, random heatmap also increases the trust level significantly, suggesting that even irrelevant details can increase the trust of humans in machine predictions.
The fact that heatmap is significantly more effective than random heatmap (78.7\% vs. 73.4\%, $p<0.001$) 
indicates that humans can interpret valuable information in weight coefficients beyond ``the placebo effect''.

\para{Humans tend to trust machine predictions more when machine predictions are correct. (\figref{fig:exp-trust-predictions}).}
We next examine whether humans trust machine predictions more when machine predictions are correct than when they are incorrect.
\figref{fig:exp-trust-predictions} shows 
that in all the five experimental setups with predicted labels, our participants trust correct machine predictions more than incorrect ones.
However, the difference is statistically significant only in {\em predicted label w/ accuracy} ($p<0.001$) and 
{\em predicted label w/ heatmap (random) ($p=0.015$)}.
These results suggest that humans can somewhat
differentiate correct machine predictions from incorrect ones.
Further evidence is required to fully understanding the reasons 
why humans (don't) trust (in)correct machine predictions.
Such understandings can improve both machine learning models and their presentations to support human decision making.

\begin{figure*}
\centering
\begin{subfigure}[t]{0.49\textwidth}
  \centering
  \includegraphics[width=0.9\textwidth]{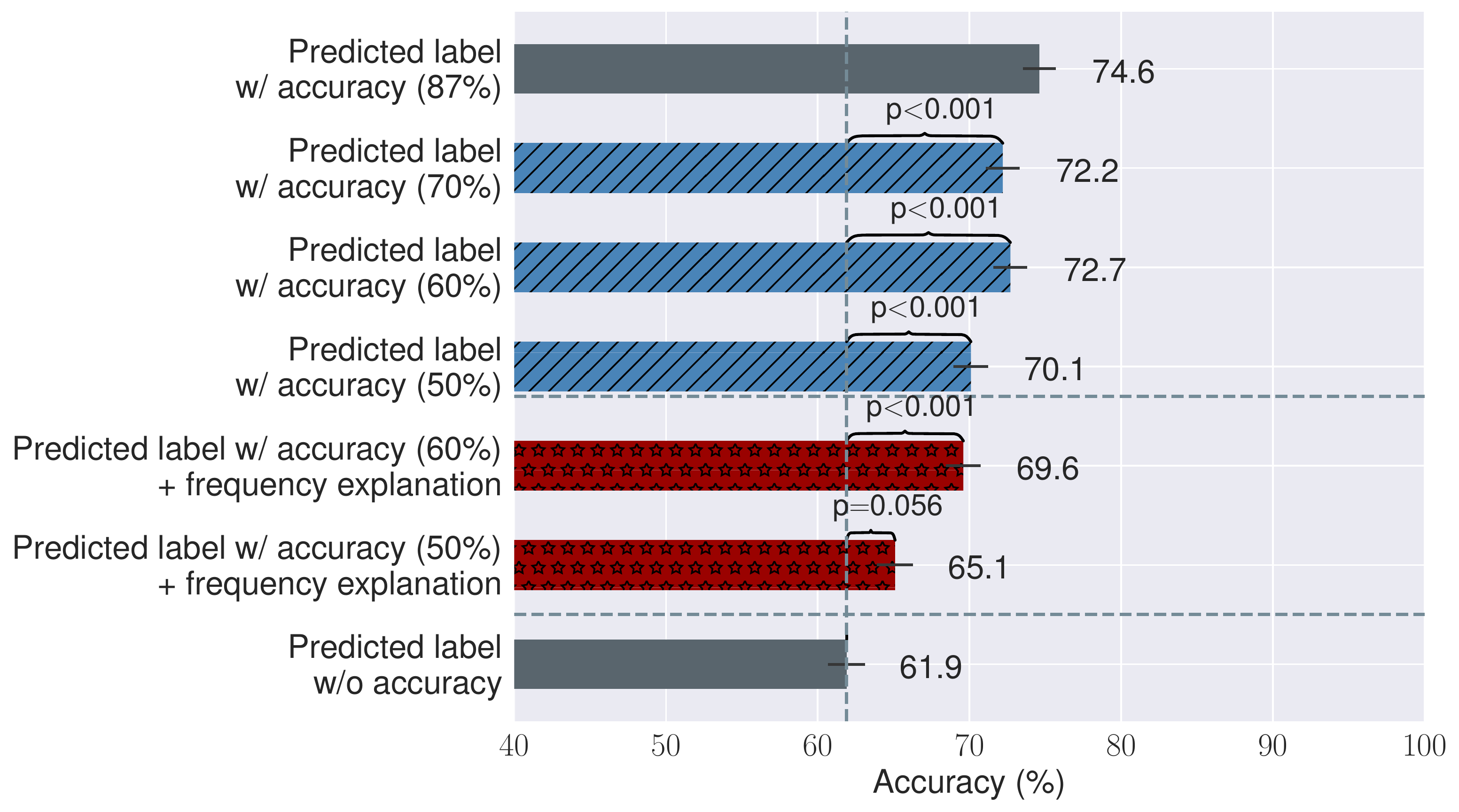}
  \caption{Human accuracy.}
  \label{fig:exp-varying-accuracy}
\end{subfigure}
\begin{subfigure}[t]{.49\textwidth}
  \centering
  \includegraphics[width=0.9\textwidth]{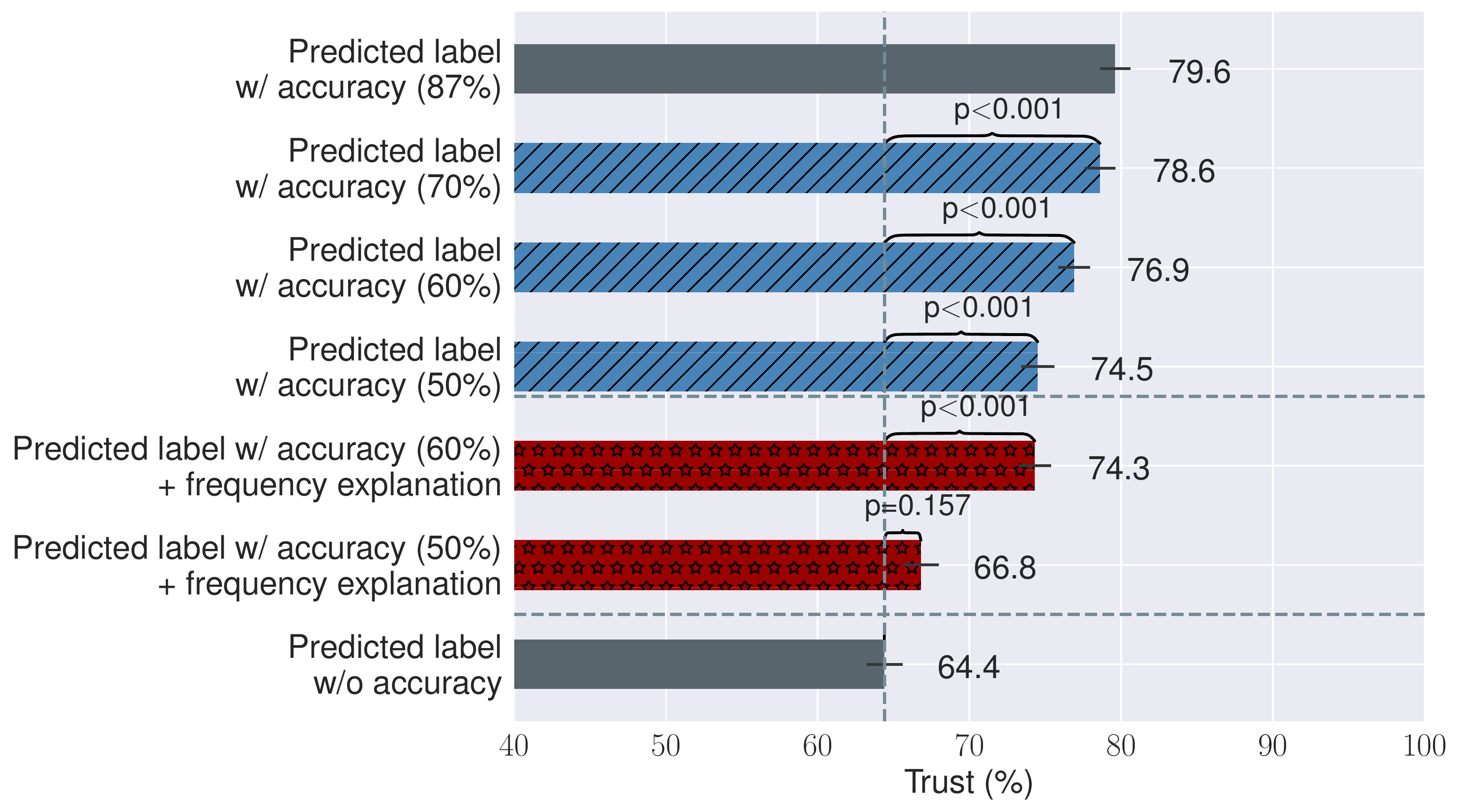}
  \caption{Trust.}
  \label{fig:exp-varying-trust}
\end{subfigure}
\caption{
Human accuracy and trust given varying statements of machine accuracy.
\figref{fig:exp-varying-accuracy} and \figref{fig:exp-varying-trust} show that human accuracy and trust generally decline with statements of decreasing machine accuracy despite the fact that machine predictions remain unchanged. 
Note that the decline of human trust with statements of decreasing accuracy is small.
Only by adding frequency explanations, human accuracy and trust become 
closer to not showing any indication of machine accuracy, i.e., {\em predicted label w/o accuracy}.
}
\end{figure*}

\subsection{Heterogeneity in Human Perception and Performance}

We finally discuss the heterogeneity between 
participants in our study.
Here we focus on the participants' estimation of their own performance and gender differences.
Refer to the appendix for additional comparisons.

\para{Human estimation of their own performance (\figref{fig:exp-estimate}).}
We ask participants to estimate their own performance in our exit survey.
Our results are not exactly aligned with the previous finding
that humans tend to overestimate their capacity of detecting lying \citep{elaad2003effects}.
In fact, $\sim$42\% of the participants correctly predicted their performance.
Among the remaining,
$\sim$18\%
 overestimated their performance, while 
 $\sim$40\%
underestimated their performance. 
\figref{fig:exp-estimate} shows the breakdown for three experimental setups.
In general, it seems difficult for humans to estimate their performance.
One participant 
who overestimated his performance (estimated 11-15 but got 10 correct) said, {\em ``I enjoyed this hit. When I was a young man, I was a manager in the hotel business and got to read a lot of comment cards from guests. I hope that I was pretty accurate in my answers''}.
Another participant who underestimated his performance (estimated 6-10 but got 15 correct)
said,  {\em ``It was difficult to determine if they were genuine or deceptive.  I don't feel certain on any of my choices''}.

\para{Heterogeneity in performance across individuals (\figref{fig:exp-demo}).} 
We have so far focused on average human performance comparisons between 
different experimental setups.
It is important to recognize that the performance of individuals can vary.
Exit survey responses allow us to study such heterogeneity.
We focus on two properties in the interest of space.
Refer to the appendix for a complete discussion of heterogeneity between individuals.

First, individuals who find the hints useful outperform those who find the hints not useful.
The difference between these two groups in Figure \ref{fig:exp-demo} ({\em predicted label + heatmap}) is statistically significant.
This observation resonates with our analysis regarding the trust of humans in machine predictions and holds in 5 out of 8 experimental setups (this question was not asked in {\em control}), although the differences are  
only statistically significant in three setups.\footnote{The low number of statistically significant differences is expected, because human performance is low unless we show predicted labels.}
Second, we find that females generally outperform males.
This observation holds in 8 out of 9 experimental setups, but none of the differences is statistically significant.
Our results 
contribute to mixed observations regarding gender differences in deception detection \cite{mann2004detecting,depaulo1993sex,mccornack1990women,li2011sex}.

\section{Varying Statements of Machine Accuracy}
\label{sec:varying_accuracy}

Given the strong influence of predicted labels and machine accuracy, a natural question to ask is how human judgment changes if we vary the statement of machine accuracy.
For example, instead of the true accuracy of 87\%, we 
could claim
that the machine has an accuracy of 60\%.
It is important to emphasize that since these statements of accuracy are not true, we do not recommend this approach as part of our \spectrum in \figref{fig:spectrum} and thus put these results in a separate section.
However, we think that it is valuable to understand how varying statements of accuracy might influence human predictions.

\para{Although human accuracy and trust generally decline with statements that suggest lower accuracy, statements of machine accuracy 
improve human trust in machine predictions even when the claimed accuracy is only 50\%.} 
To understand human accuracy with varying statements of machine accuracy, we use {\em predicted label w/o accuracy} and {\em predicted label w/ accuracy (87\%)} as benchmarks.
In \figref{fig:exp-varying-accuracy} and \figref{fig:exp-varying-trust},
human accuracy and trust with varying statements of machine accuracy all fall between these two benchmarks as expected.
Here we focus on the blue bars filled with forward slashes that correspond to simple statements of machine accuracy, ``The machine predicts that the below review is deceptive. It has an accuracy of approximately $x$\%'' ($x = 70, 60, 50$).
As the claimed accuracy declines from 87\% to 50\%, human accuracy and trust
decrease, with the exception of human accuracy from 70\% to 60\%.
However, the decline in human trust and accuracy is fairly small.
For instance, {\em predicted label w/ accuracy (50\%)} still outperforms {\em predicted label w/o accuracy} significantly.
The results are surprising and counterintuitive since one should put less trust in a machine that has only an accuracy of 50\% as compared to a machine that boasts 87\%.
Our findings suggest that any indication of machine accuracy, be it high or low, improves human trust in the machine.
This observation echoes prior work on numeracy that suggests that average humans and even doctors struggle with interpreting and acting on numbers \citep{reyna2008numeracy,berwick1981doctors,slovic2006risk,peters2006numeracy}.
Therefore, it is crucial that we develop a better {\em empirical} understanding of how humans interact with explanations and predictions of machine learning models in decision making before using these machine learning models in the loop of human decision making.

\para{Frequency explanations can help humans 
interpret and act on statements of machine accuracy.} 
To further investigate human interaction with varying statements of machine accuracy, we add frequency 
explanations to the statement with accuracy 50\% and 60\%.
Specifically, we show participants ``The machine predicts that the below review is {\em deceptive}. It has an accuracy of approximately 50\%, which means that it is correct 5 out of 10 times.'' instead of ``The machine predicts that the below review is {\em deceptive}. It has an accuracy of approximately 50\%.''
The results are shown with the red bars filled with stars in \figref{fig:exp-varying-accuracy} and \figref{fig:exp-varying-trust}.
We find that frequency 
explanations reduce the trust that humans place on machine predictions.
For instance, human accuracy in {\em predicted label w/ accuracy (50\%) + frequency explanation} is $\sim$7\% lower ({\em p=0.003}) than in {\em predicted label w/ accuracy (50\%)}.
Similarly, human trust in {\em predicted label w/ accuracy (50\%) + frequency explanation} is $\sim$10\% lower ({\em p<0.001}) than in {\em predicted label w/ accuracy (50\%)}.
Furthermore, the differences in human accuracy and trust are not statistically significant between {\em predicted label w/ accuracy (50\%) + frequency explanation} and {\em predicted label w/o accuracy}.
These observations suggest that frequency 
explanations can help humans interpret
statements of machine accuracy, in which case a statement of 50\% accuracy with frequency 
explanation is almost the same as not showing machine accuracy.
Our frequency explanations are also known as frequent format 
and have been shown to be more effective for conveying uncertainty than stating the probability \citep{sedlmeier2001teaching,gigerenzer1996psychology,gigerenzer1995improve}.

\section{Concluding Discussion}
\label{sec:conclusion}

In this paper, we conduct the first empirical study to investigate whether machine predictions and their explanations
can improve human performance in challenging tasks such as deception detection.
We propose a \spectrum between full human agency and full automation, and design machine assistance with varying levels of priming along the \spectrum.
We find that explanations alone 
slighlty improve human performance, while showing predicted labels 
significantly improves human performance.
Adding an explicit statement of strong machine performance can further improve human performance.
Our results demonstrate a tradeoff between human performance and human agency, and explaining machine predictions may moderate this tradeoff.

We find interesting results regarding the trust that humans place on machine predictions.
On the one hand, humans tend to trust correct machine predictions more than incorrect ones, which indicates that it is possible to improve human decision making while retaining human agency.
On the other hand, we show that human trust can be easily enhanced by adding random heatmap as explanations or statements of low accuracies that do not justify trusting machine predictions.
In other words, additional details including irrelevant ones can improve the trust that humans place on machine predictions.
These findings highlight the importance of taking caution in using machine learning for supporting decision making and developing methods to improve the transparency of machine learning models and its associated human interpretation.

As machine learning gets employed to support decision making in our society, it is crucial that the machine learning community not only advances machine learning models, but also develops a better understanding of how these machine learning models are used and how humans interact with these models in the process of decision making.
Our study takes an initial step towards understanding human predictions with assistance from machine learning models in challenging tasks.

\para{Implications and future directions.} 
Our results show that explanations alone slightly improve human performance.
One reason for the limited improvement with explanations alone is that although we provide explanations during the decision making process, we provide limited resources to ``teach'' these explanations.
A possible future direction is to develop tutorials for machine learning models and their explanations to relieve some cognitive burden from humans, e.g., summarizing the model as a list of rules, adding heatmap in examples or providing a sequence of training examples with explanations and sufficient coverage.
This direction also connects to the area of machine teaching \citep{zhu2015machine,mac2018teaching,singla2014near}.

Another possible direction to improve the effectiveness of explanations is to provide narratives. 
Our results suggest that feature-based and example-based explanations provide useful details for machine predictions to improve the trust of humans in machine predictions.
It can be useful if we can similarly provide rationales behind feature-based and example-based explanations in the form of narratives.
A qualitative understanding of how turkers interpret hints from machine learning models
may shed light on the requirements of effective narratives.

Last but not least, it is important to study the ethical concerns of providing assistance from machine learning models in human decision making.
Our results demonstrate a clear tradeoff in this space: it is difficult to improve human performance without showing predicted labels, but showing predicted labels, especially alongside machine performance, runs the risk of removing human agency.
Human decision makers with assistance from machines
further complicate the current discussions on the issue of fairness in algorithmic decision making~\citep{kleinberg2016inherent,hardt2016equality,corbett2017algorithmic}.
As the adoption of machine learning approaches can have broad impacts on our society, such questions require inputs from machine learning researchers, legal scholars, and the entire society.

\para{Limitations.} 
We use Amazon Mechanical Turk to recruit participants, but this may not be a representative sample of the population.
However, we would like to emphasize that turkers are likely to provide a better proxy than machine learning experts for understanding how humans interact with assistance from machine learning models in critical challenging tasks.
Also, our explanations are derived from a linear SVM classifier and nearest neighbors.
It may be even more challenging for humans to interpret explanations of non-linear classifiers.

Another important challenge in understanding how humans interact with machine learning models lies in the difficulty to assess the generalizability of our results.
Our formulation of deception detection represents a scenario where machines outperform humans by a wide margin and humans may have developed false beliefs about this task, as most humans have read reviews online.
In order to consider a wide range of tasks, e.g., bail decisions and medical diagnosis, we need a framework to compare different tasks.
Machine performance and humans' prior intuition are probably important factors that can influence human interpretation of the explanations.
However, it remains an open question whether there exists a principled framework to reason about these tasks.
At the very least, it is important for our community to go beyond simple visual tasks such as OCR and object recognition, especially for the purpose of improving human performance in decision making.

\bibliographystyle{ACM-Reference-Format}
\bibliography{refs}

\begin{acks}
We would like to thank Elizabeth Bradley, Michael Mozer, Sendhil Mullainathan, Amit Sharma, Adith Swaminathan, and anonymous reviewers for helpful discussions and feedback. This material is based upon work supported by the National Science Foundation under Grant No. 1837986.
Any opinions, findings, and conclusions or recommendations expressed in this material are those of the author(s) and do not necessarily reflect the views of the National Science Foundation.
\end{acks}

\appendix
\section{Appendix}
\label{sec:appendix}

\subsection{Amazon Mechanical Turk Setup}
To ensure quality results, we include several criteria for turkers:
1) the turker is based in the United States so that we assume English fluency;
2) the turker has completed at least 50 HITs (human intelligence tasks);
3) the turker has an approval rate of at least 99\%.

\begin{figure}[h]
  \centering
  \includegraphics[width=0.4\textwidth]{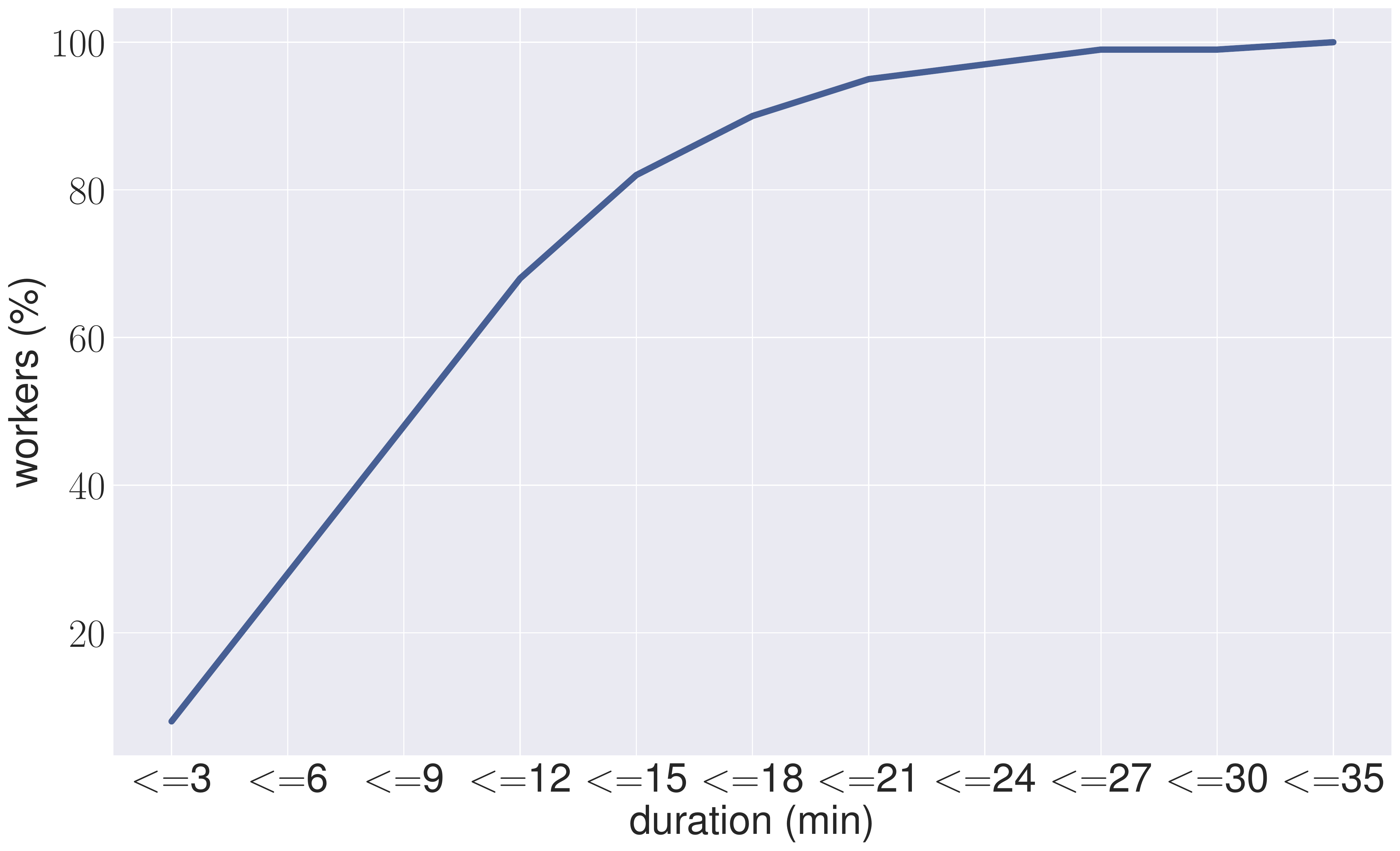}
  \caption{Cumulative distribution of study duration in 9 experimental setups.
  }
  \label{fig:time-distribution}
\end{figure}

\begin{figure}[h]
  \centering
  \includegraphics[width=0.4\textwidth]{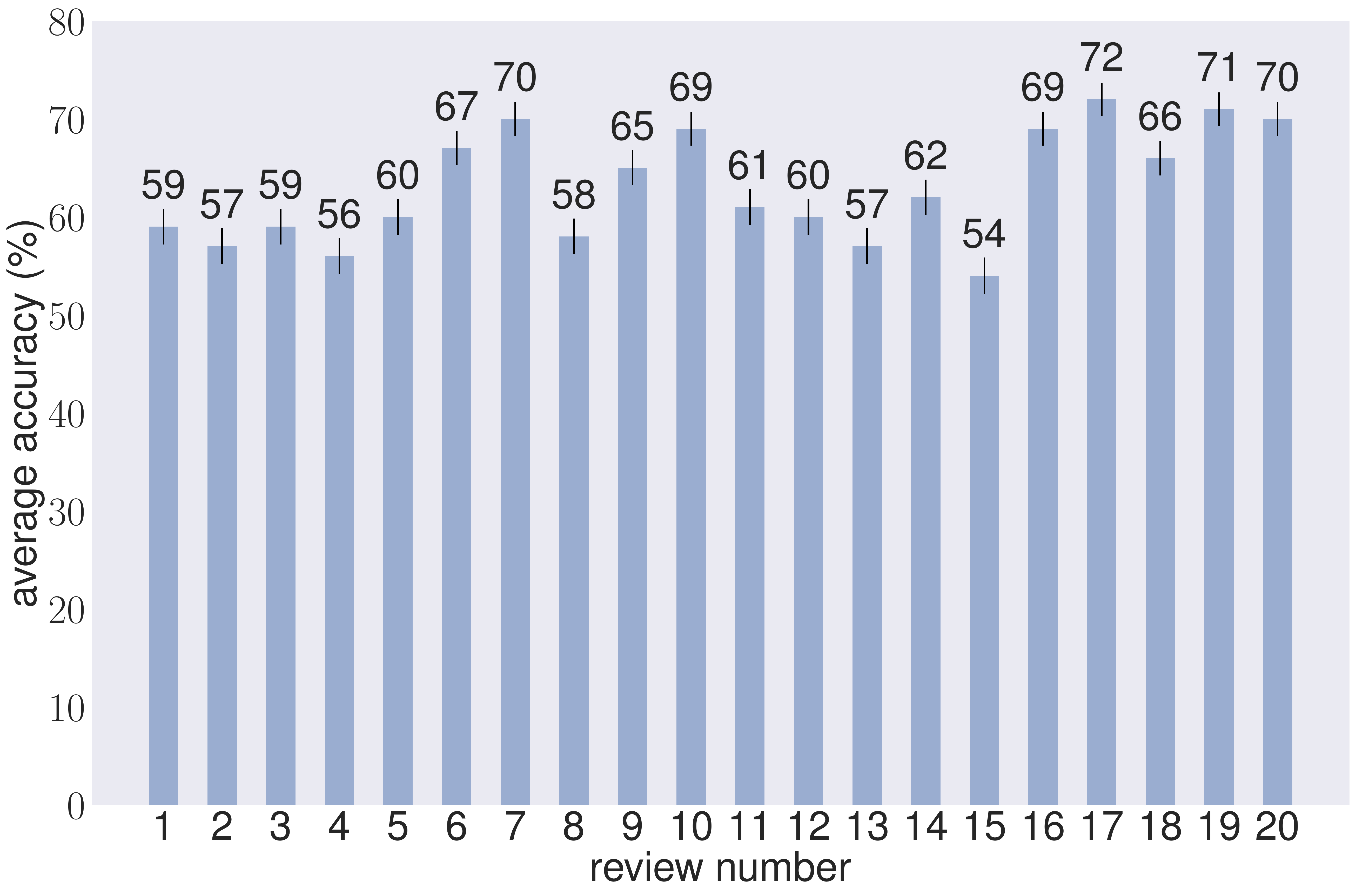}
  \caption{Average accuracy with respect to review ordering in 9 experimental setups.}
  \label{fig:accuracy-per-review}
\end{figure}

\begin{figure*}[t]
 \centering
  \includegraphics[width=0.9\textwidth]{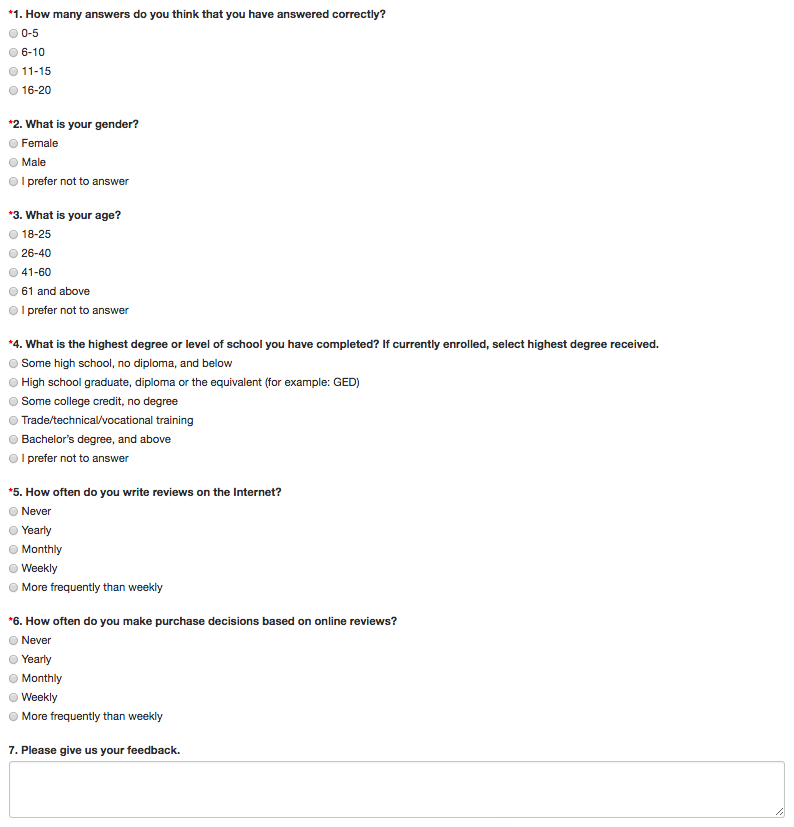}
  \caption{Survey questions for control group.}
  \label{fig:exp-survey-control}
\end{figure*}

\begin{figure*}[t]
  \centering
  \includegraphics[width=0.9\textwidth]{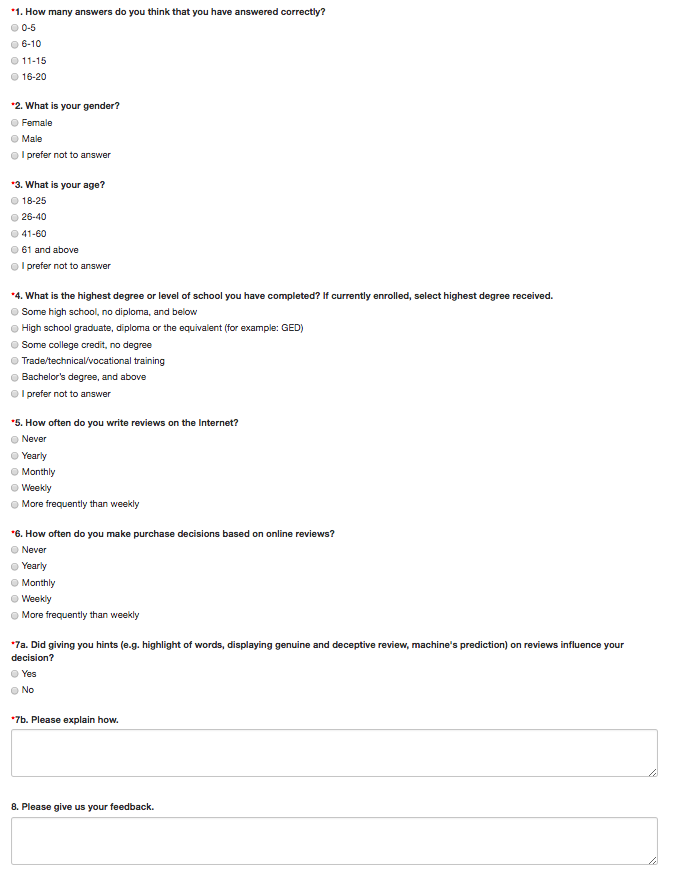}
  \caption{Survey questions for all the other groups.}
  \label{fig:exp-survey-config}
\end{figure*}

Before working on the main task, turkers need to go through a short training session, in which we show three reviews from the training data.
We present the correct answer after turkers make their prediction.
The interface during training is exactly the same as in the actual experiment.
After making predictions for 20 reviews, turkers are required to fill out an exit survey that solicits their estimation of their own performance in this task and basic demographic information including age, gender, education background, and experience with online reviews (screenshots in \figref{fig:exp-survey-control} and \figref{fig:exp-survey-config}).
If the HIT is approved, the turker is compensated a dollar and bonuses depending on the number of reviews he correctly predicted.
For example, if a turker makes 11 correct predictions, he is compensated \$0.22 in addition to a dollar.
The average duration for finishing our HIT is about 11 minutes (\figref{fig:time-distribution} shows 
the CDF of 
the duration).
Turkers spend the shortest amount of time on average (8.3 minutes) in {\em predicted labels w/ accuracy} and the longest amount of time on average (14.4 minutes) in {\em examples}, which is consistent with our expectation about extra cognitive burden from reading two more reviews.
To sanity check that participants pay similar attention throughout the study, \figref{fig:accuracy-per-review} shows the average accuracy with respect to the order in which reviews show up\footnote{Thanks to suggestions from anonymous reviewers.}: there does not exist a downward trend.
All results are based on the 9 experimental setups in Section 4 of the main paper and results with varying statements of accuracy are not included.

\subsection{Experiment Interfaces}
This section shows example interfaces for the other five experimental setups that are not shown in the main paper ({\em predicted label + heatmap (random)} has the same interface as {\em predicted label + heatmap} except that words are highlighted randomly).

\begin{figure*}[t]
\centering
    \begin{subfigure}[t]{.9\textwidth}
    \centering
    \includegraphics[width=\textwidth]{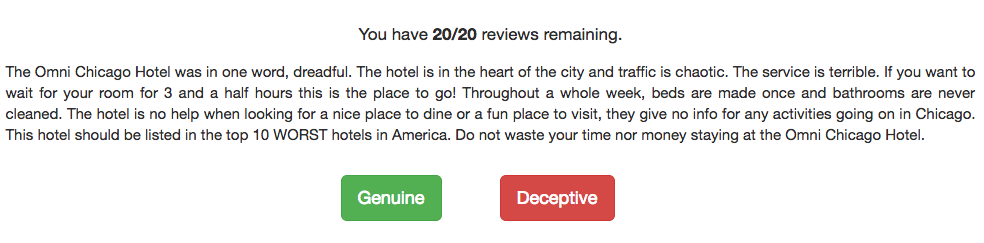}
    \caption{Example interface for {\em control}.}
    \label{fig:exp-control}
    \end{subfigure}
    \begin{subfigure}[t]{.9\textwidth}
    \centering
    \includegraphics[width=\textwidth]{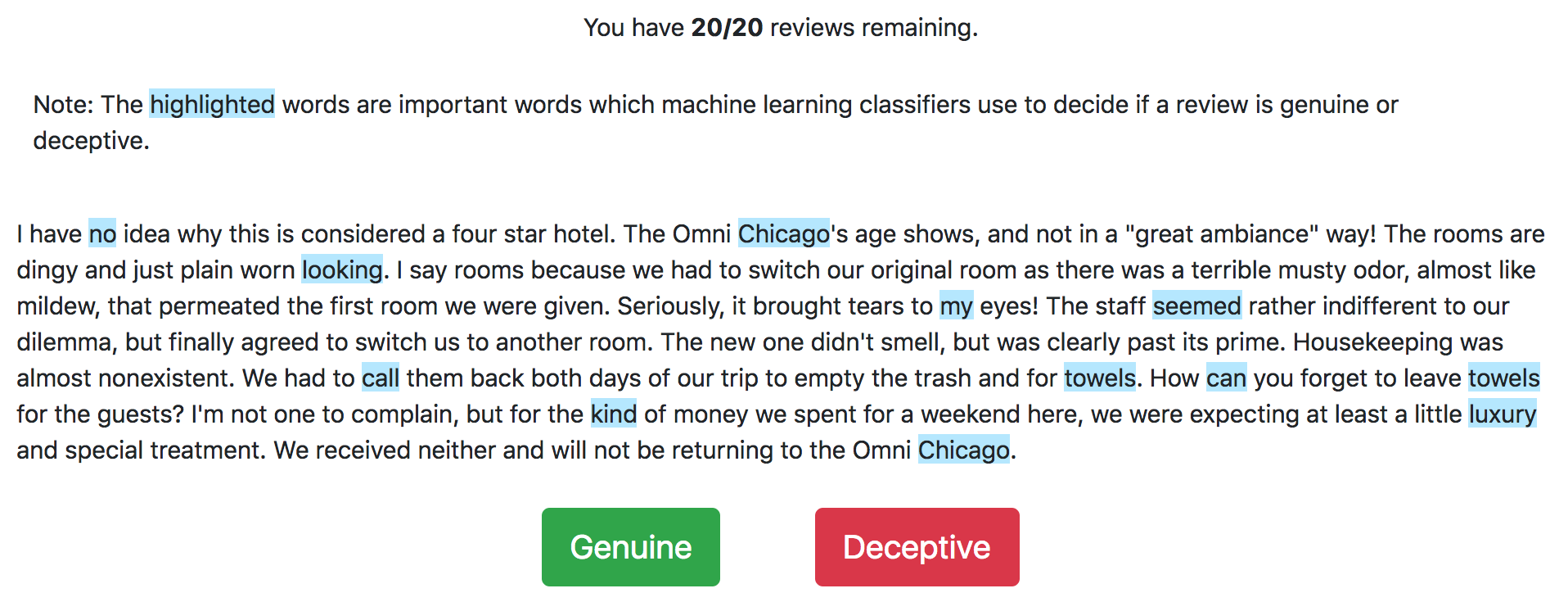}
    \caption{Example interface for {\em highlight}.}
    \label{fig:exp-highlight}
    \end{subfigure}
\end{figure*}

\begin{figure*}[t]
\centering
    \begin{subfigure}[t]{.9\textwidth}
    \centering
    \includegraphics[width=\textwidth]{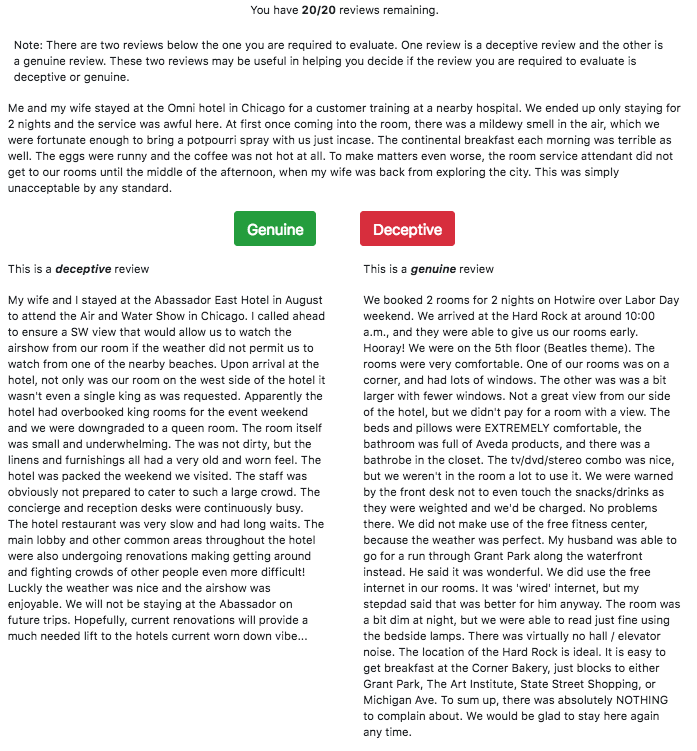}
    \caption{Example interface for {\em examples}.}
    \label{fig:exp-display}
    \end{subfigure}
    \begin{subfigure}[t]{.9\textwidth}
    \centering
    \includegraphics[width=\textwidth]{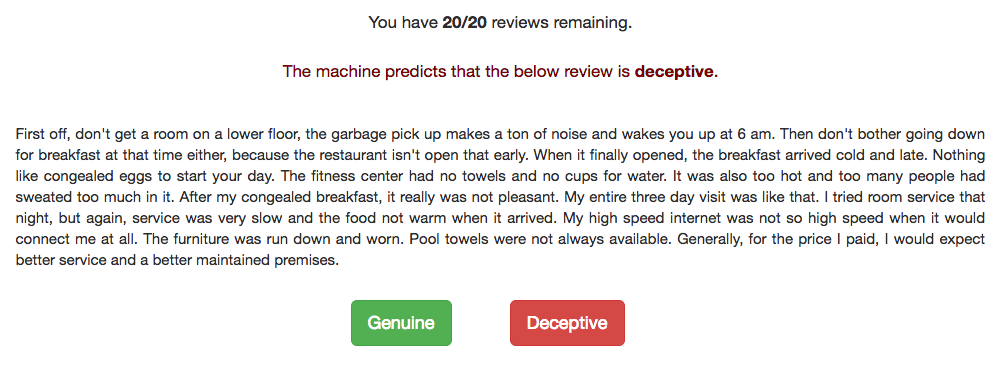}
    \caption{Example interface for {\em predicted label w/o accuracy}.}
    \label{fig:exp-wo-acc}
    \end{subfigure}
\end{figure*}

\begin{figure*}[t]
    \begin{center}
    \includegraphics[width=0.9\textwidth]{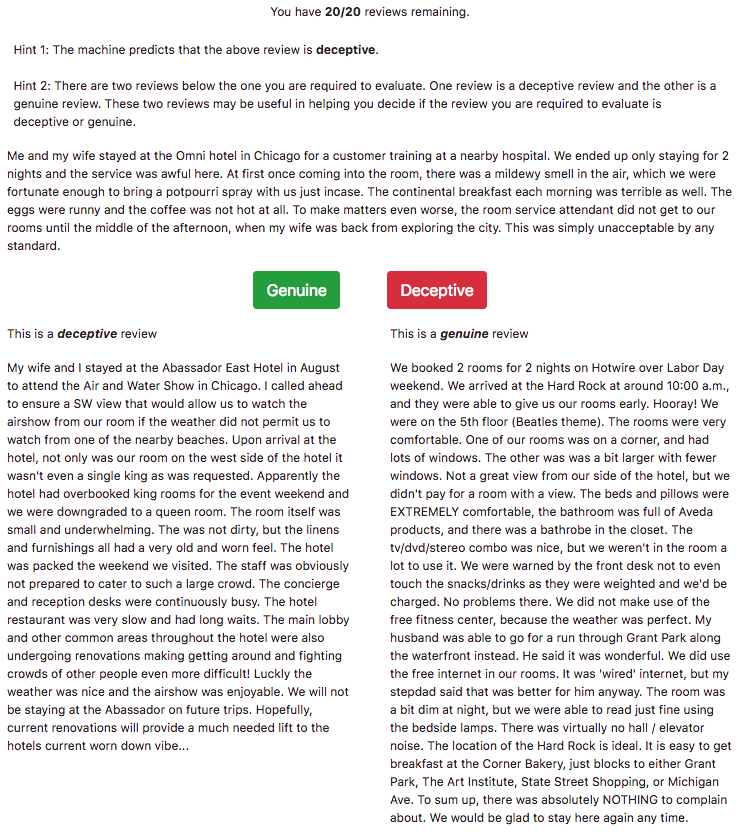}
    \end{center}
    \caption{Example interface for {\em predicted label + examples}.}
\label{fig:exp-combi2}
\end{figure*}

\begin{itemize}
  \item Control (\figref{fig:exp-control}).
  \item Highlight (\figref{fig:exp-highlight}).
  \item Examples (\figref{fig:exp-display}).
  \item Predicted label w/o accuracy (\figref{fig:exp-wo-acc}).
  \item Predicted label + examples (\figref{fig:exp-combi2}).
\end{itemize}

\subsection{Individual Differences}
Here we present further results on heterogeneous performance among individuals. 
We present figures for four experimental setups that are representative of different levels of priming: {\em heatmap}, {\em examples}, {\em predicted label w/o accuracy}, and {\em predicted label + heatmap}.

\begin{figure}[h]
  \centering
  \includegraphics[width=0.45\textwidth]{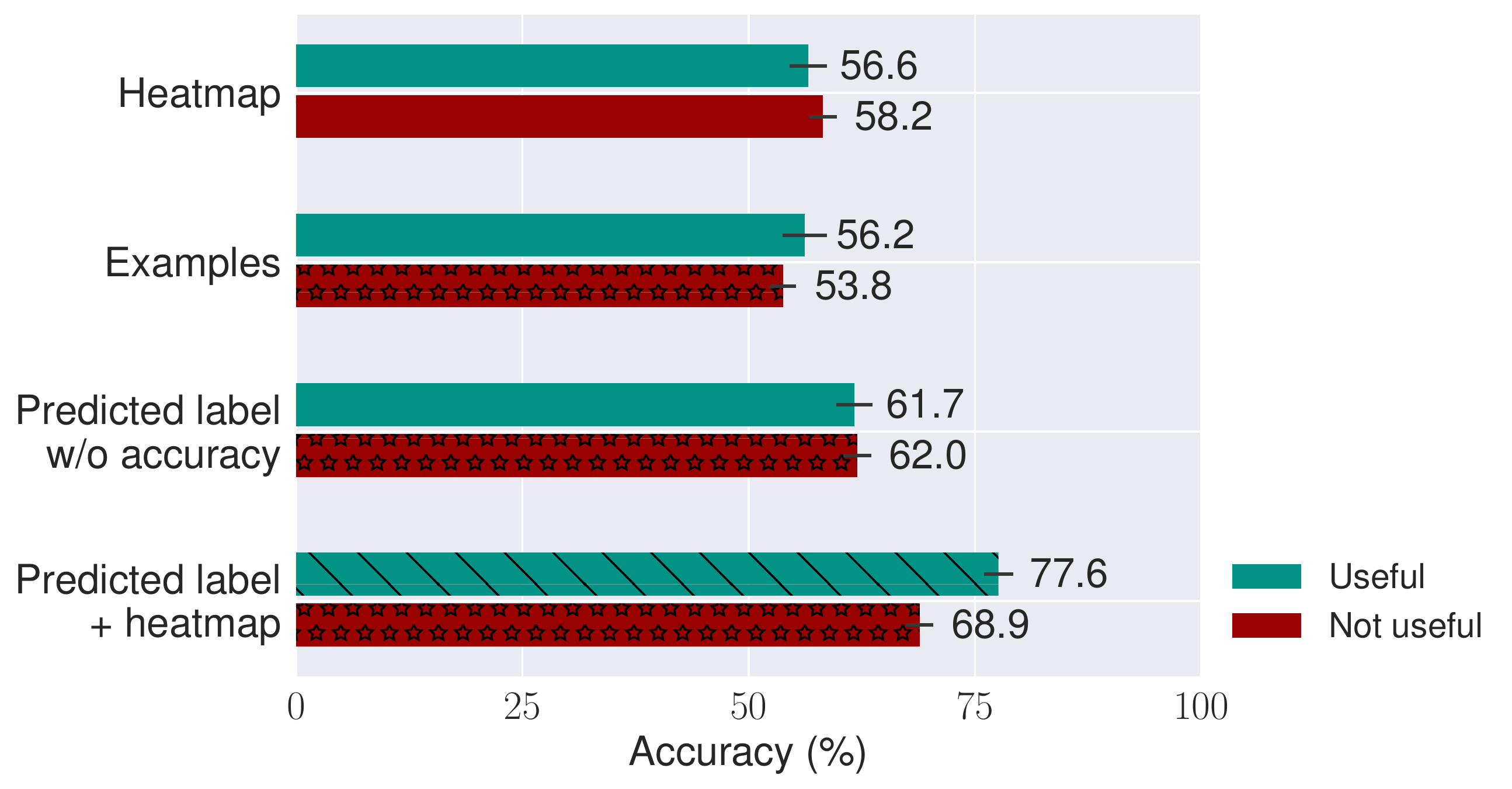}
  \caption{Human accuracy vs. usefulness of hints.}
  \label{fig:het-usefulness}
\end{figure}

\para{Hint usefulness (\figref{fig:het-usefulness}).} 
As discussed in the main paper, human performance is better for participants who find hints useful than those who do not find hints useful in 5 out of 8 experimental setups. 
{\em Highlight}, {\em heatmap} and {\em predicted label w/o accuracy} are the exceptions.
The difference in three setups ({\em predicted label + heatmap}, {\em predicted label + heatmap (random)}, {\em predicted label w/ accuracy}) is statistically significant.

\begin{figure}[h]
  \centering
  \includegraphics[width=0.45\textwidth]{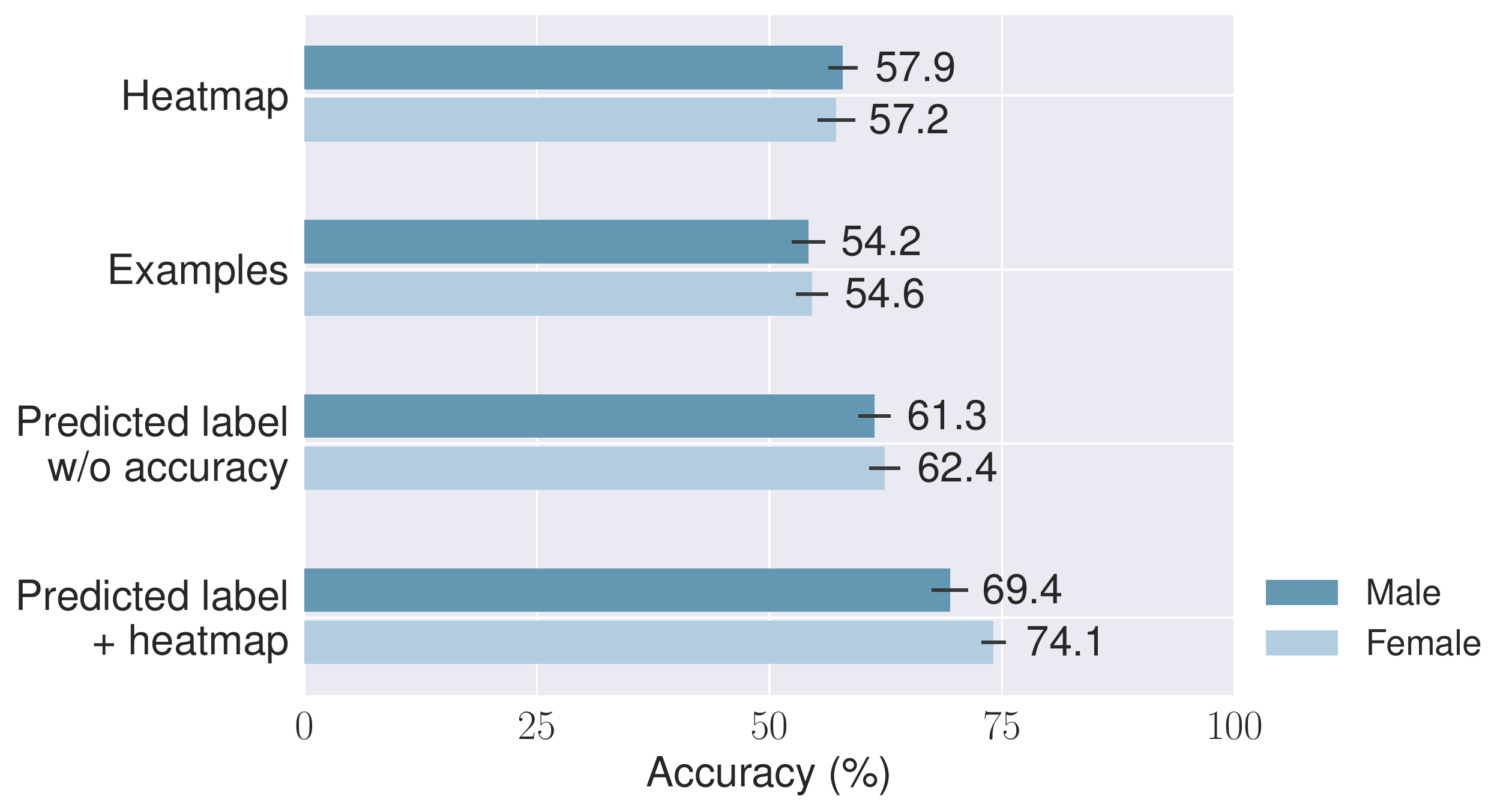}
  \caption{Human accuracy vs. gender.}
  \label{fig:het-gender}
\end{figure}

\para{Gender differences (\figref{fig:het-gender}).} 
Females generally outperform males, in 8 out of 9 experimental setups.
None of the differences is statistically significant.

\begin{figure}[h]
  \centering
  \includegraphics[width=0.45\textwidth]{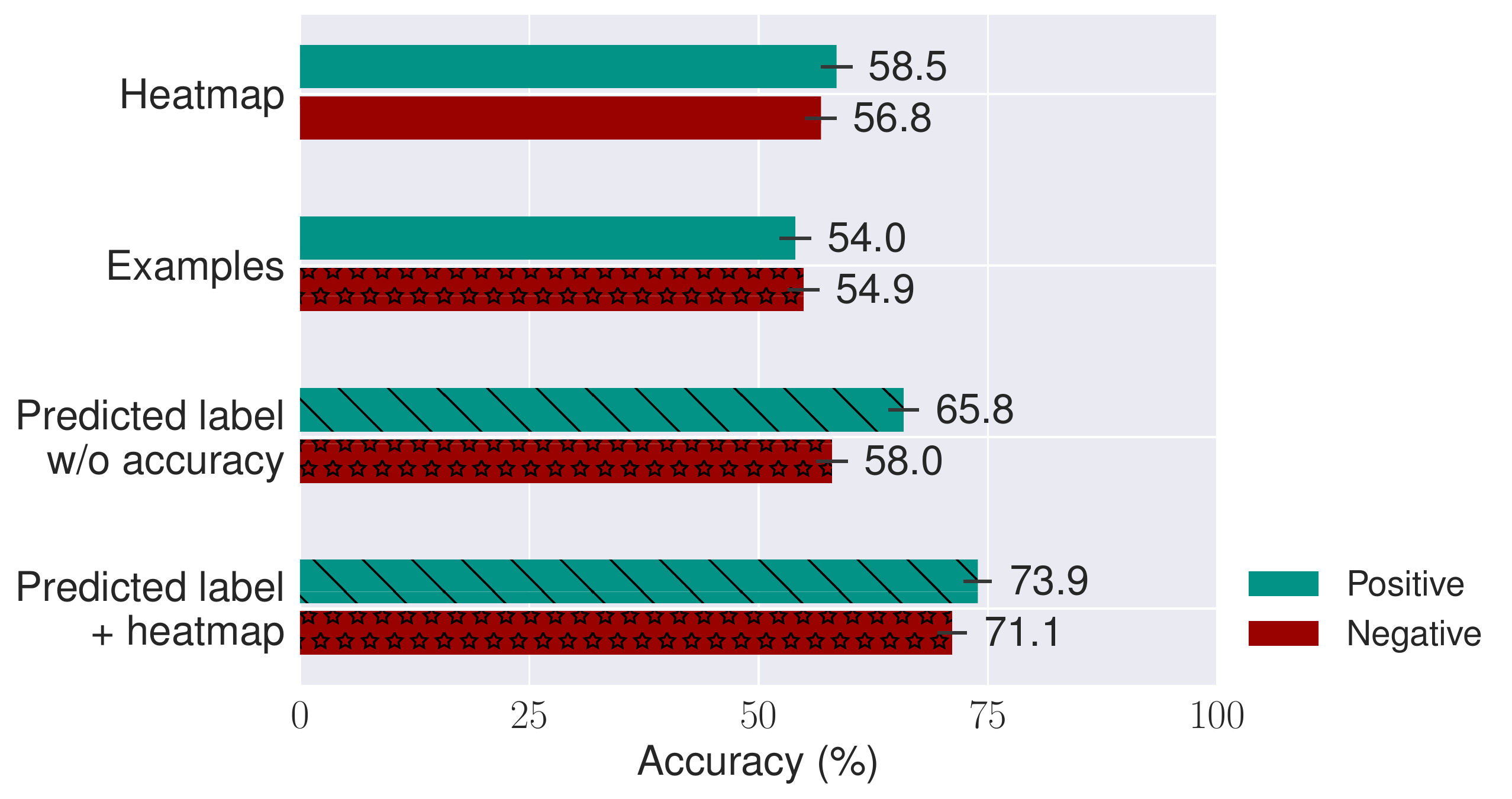}
  \caption{Human accuracy vs. review sentiment.}
  \label{fig:het-sentiment}
\end{figure}

\para{Review sentiments (\figref{fig:het-sentiment}).}
One possible hypothesis is that humans perform differently depending on the sentiment of reviews.
Indeed, we observe that humans consistently perform better for positive reviews (8 out 9 experimental setups).
However, the difference is only statistically significant for {\em predicted label w/o accuracy}.

\begin{figure}[h]
  \centering
  \includegraphics[width=0.44\textwidth]{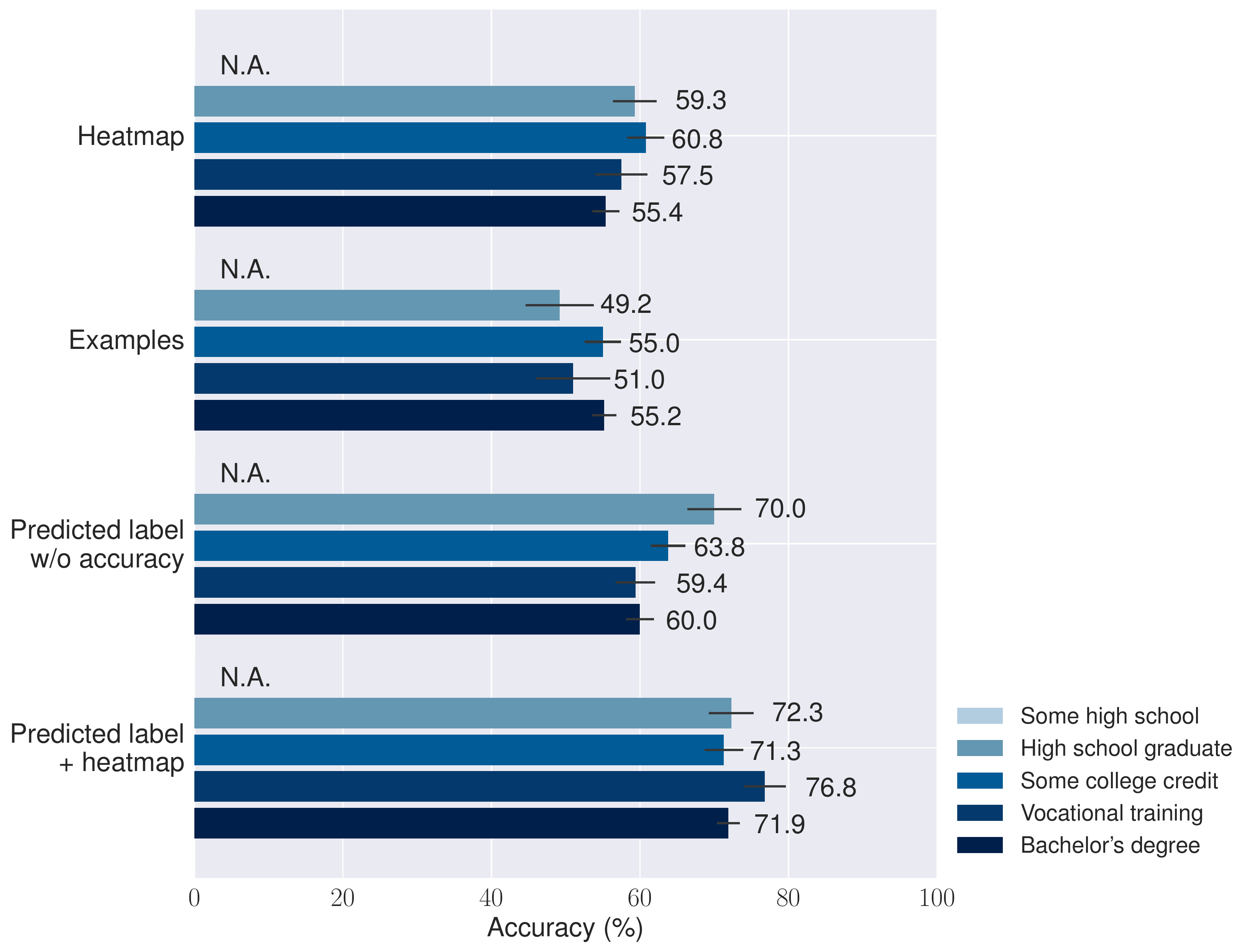}
  \caption{Human accuracy vs. education background.}
  \label{fig:het-edu}
\end{figure}

\para{Education background (\figref{fig:het-edu}).} There is no clear trend regarding education background, which suggests that education levels do not correlate with the ability to detect deception.
For instance, high school graduates perform the best in {\em predicted label w/o accurcay}, but the worst in {\em examples}.
Since there are five groups, each group is relatively sparse.
We thus did not conduct statistical testing for these observations.

\begin{figure}[h]
  \centering
  \includegraphics[width=0.44\textwidth]{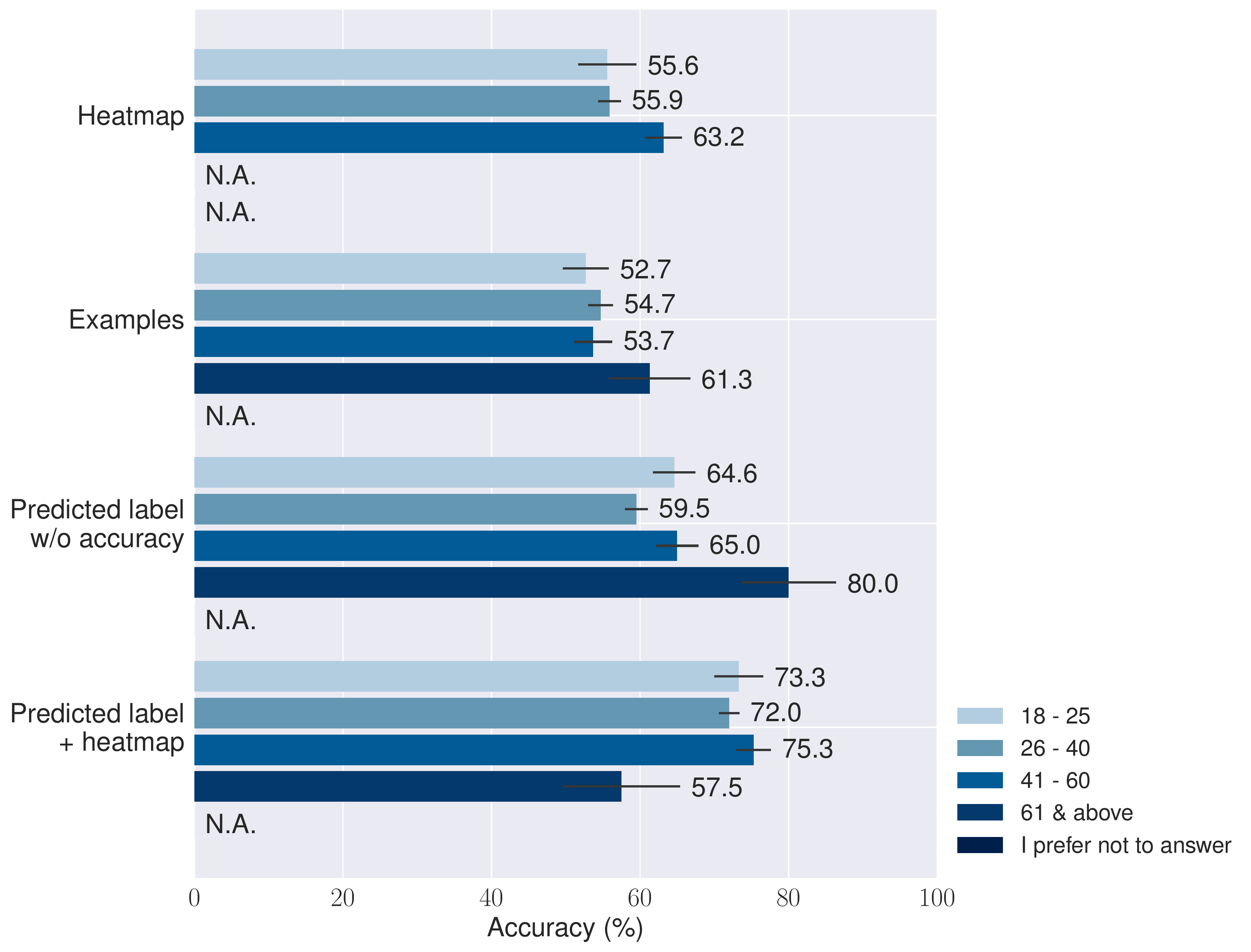}
  \caption{Human accuracy vs. age groups.}
  \label{fig:het-age}
\end{figure}

\para{Age group (\figref{fig:het-age}).} There is no clear trend regarding age groups either.
For instance, participants that are 61 \& above perform the best in {\em predicted label w/o accuracy}, but worst in {\em predicted label + heatmap}.
Similarly, since there are five groups and that each group is also relatively sparse, we did not conduct statistical testing for these observations.

\begin{figure}[h]
  \centering
  \includegraphics[width=0.44\textwidth]{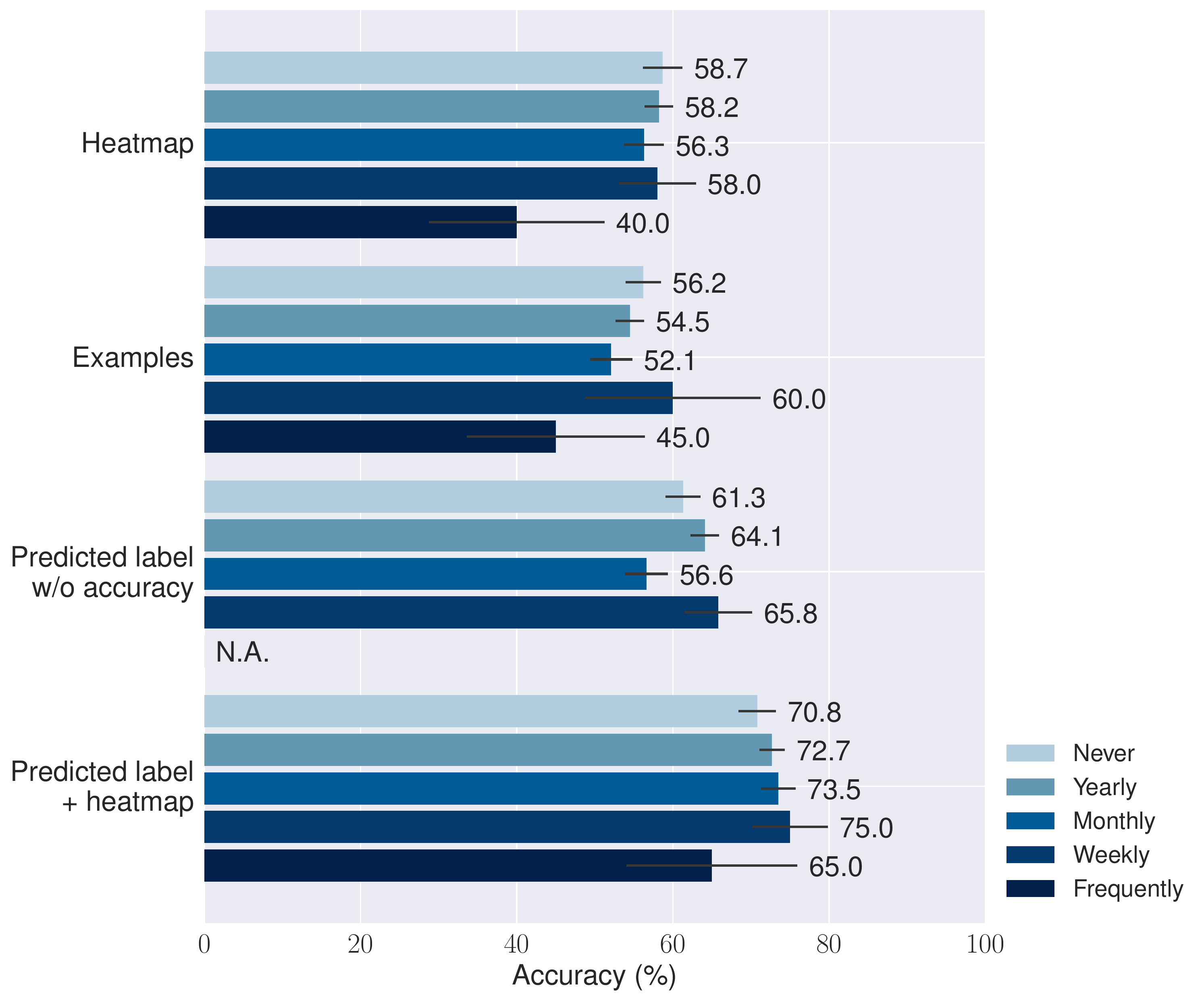}
  \caption{Human accuracy vs. review writing experience.}
  \label{fig:het-reviews}
\end{figure}

\para{Review experience (\figref{fig:het-reviews}).}
There is no clear trend regarding experience of writing reviews.
With the exception of {\em control} and {\em predicted label + heatmap (random)}, 
the group that reports the best performance is either users who write reviews weekly or users who write reviews frequently.
Again, we did not conduct statistical testing for review experience.

\end{document}